%% file: main.tex
\documentclass[10pt]{article}

\usepackage{style/arxiv}
\usepackage{lipsum}
\usepackage{tikz}
\usepackage{fontawesome5}
\usepackage{tcolorbox}
\usepackage{pifont}
\usepackage{graphicx}
\usepackage{subcaption}
\usepackage{amsmath}
\usepackage{booktabs}
\usepackage{tikz}
\usepackage{fontawesome5}
\usepackage{xspace}
\usepackage{tcolorbox}
\usepackage{float}
\usepackage[ruled,vlined,linesnumbered]{algorithm2e}
\usepackage{amsmath}

\definecolor{blockcolor}{gray}{0.92}
\definecolor{lightred}{HTML}{f6c6ad}
\definecolor{lightgreen}{HTML}{b4e5a2}
\definecolor{lightblue}{HTML}{a6caec}
\definecolor{Ablue}{HTML}{C5DCF1}
\definecolor{Agreen}{HTML}{D5EBEA}

\newcommand{\name}{\textbf{DIG}}
\newcommand{\keyframe}{\textit{r-frame}}
\newcommand{\keyframes}{\textit{r-frames}}
\newcolumntype{Z}{>{\centering\arraybackslash}p{1.1cm}}

\begin{document}

\shorttitle{Adapting Frame Selection to Query Types for Long-Form Video Understanding}
\shortauthor{Li \etal}
\headerright{\textit{CVPR 2026}}
\firstpagefootnote{* Work done during Jialuo's internship at MSRA.}

\title{\underline{D}\underline{I}vide, then \underline{G}round: Adapting Frame Selection to Query Types for Long-Form Video Understanding}
\author{
  Jialuo Li$^{1,2,*}$ \quad
  Bin Li$^{2}$ \quad
  Jiahao Li$^{2}$ \quad
  Yan Lu$^{2}$ \\[0.3cm]
  $^{1}$Tsinghua University \quad $^{2}$Microsoft Research Asia
}
\date{}
\maketitle

\begin{abstract}
The application of Large Multimodal Models (LMMs) to long-form video understanding is constrained by limited context lengths and the computationally prohibitive cost of processing dense video tokens. Consequently, recent research has focused on query-aware frame selection, methods that often incur significant computational overhead. This paper challenges the assumption that such complex search mechanisms are universally necessary. We first identify and validate a query typology distinguishing between \textbf{global query} and \textbf{localized query}. We demonstrate that while uniform sampling is both effective and efficient for global queries, localized queries indeed necessitate query-aware selection for optimal performance. Building on this insight, we propose~\name, a training-free frame selection framework that adapts its strategy based on the query type. Specifically,~\name~employs efficient uniform sampling for global queries while activating a specialized pipeline to extract query-relevant frames for localized queries. Experiments on three long-form video understanding benchmarks demonstrate that~\name~consistently outperforms existing baselines and robustly improves LMM performance, even when scaling the input frame count to 256. The code is available at \href{https://github.com/Jialuo-Li/DIG}{{\color{LabBlue} \textit{https://github.com/Jialuo-Li/DIG}}}.
\end{abstract}

\input{sections/introduction}
\input{sections/related_work}
\input{sections/revisit}

\input{sections/method}
\input{sections/experiments}
\input{sections/discussion}
\input{sections/conclusion}

\bibliographystyle{unsrtnat}
\bibliography{references}

\newpage
\appendix
\onecolumn

\makeappendixtitle  
\input{sections/appendix}

\end{document}

%% file: sections/introduction.tex
\section{Introduction}

In recent years, there has been a rapid advancement in large multimodal models~(LMMs)~\cite{liu2023visualinstructiontuning, tong2024cambrian1fullyopenvisioncentric, li2024llavaonevisioneasyvisualtask, shi2024eagle, chen2024internvl} for open-world visual understanding. A natural and increasingly important direction within this field is the extension of these models to handle video data, thereby enabling them to perform complex video understanding tasks~\cite{maaz2023video,lin2023mm,zhang2023simple,lin2023video,jin2024chat,ren2024timechat, zhang2025videollama3frontiermultimodal, chengVideoLLaMA2Advancing2024, zhang2024videoinstructiontuningsynthetic}. The common approach~\cite{zhang2023videollamainstructiontunedaudiovisuallanguage, bai2025qwen25vltechnicalreport} involves representing videos as sequences of individual frames, where visual features are extracted from each frame and concatenated to form a video representation that is subsequently processed by the large language model~(LLM). However, due to the limited context length of the LLM and the sheer volume of video tokens, it is impractical to input all frames directly. As a result, only a sampled subset of frames is typically used as input. The predominant method is uniform sampling which selects frames at fixed intervals. While this maximizes temporal coverage, it is query-agnostic, often selecting redundant frames while omitting crucial, query-relevant moments that are essential for accurate reasoning.

To address this limitation, recent work has introduced query-aware adaptive frame selection mechanisms~\cite{liu2025boltboostlargevisionlanguage, tang2025adaptivekeyframesamplinglong, yuFrameVoyagerLearningQuery2024, ye2025rethinkingtemporalsearchlongform, sun2025mdp3trainingfreeapproachlistwise}. These methods identify and utilize the most representative frames as input based on the query, but at the cost of significant computational overhead for searching within the video. This high cost motivates a critical question that is frequently overlooked: \textit{Is such a complex search mechanism strictly necessary for all query types?} Our findings indicate that the answer is negative. We first identify the existence of two distinct query categories: \textit{global query}, which requires holistic video understanding, and \textit{localized query}, which targets specific temporal segments. We observe a significant performance disparity in uniform sampling between these categories. As the number of sampled frames increases, performance on localized queries degrades substantially, as irrelevant frames are injected into the context. Conversely, performance on global queries remains stable. This finding validates our query typology. Based on this, our further experiments demonstrate that for global queries, uniform sampling already achieves robust performance. In such cases, deploying more complex selection methods is often inefficient and yields diminishing returns. Conversely, it is for localized queries that advanced, query-aware selection mechanisms are truly impactful, delivering substantial performance gains where uniform sampling fails, highlighting the need for a dynamic, query-dependent sampling strategy.

\begin{wrapfigure}{r}{0.5\linewidth}
\centering
\vspace{-12pt}
\includegraphics[width=\linewidth]{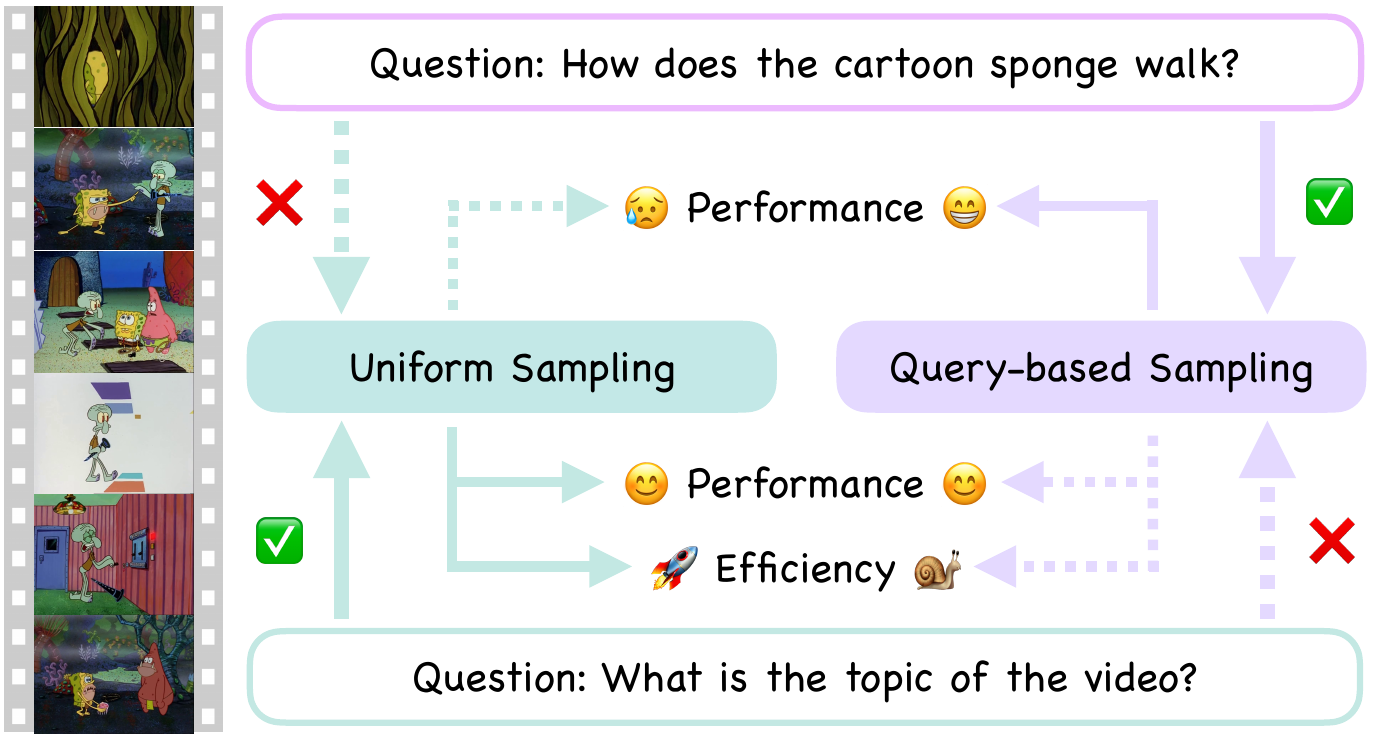}
\caption{For global queries requiring holistic understanding (bottom), uniform sampling is both effective and efficient. Conversely, for localized queries targeting specific temporal segments (top), query-based sampling is necessary to ensure high performance.}
\vspace{-12pt}
\label{fig:case}
\end{wrapfigure}

Building on these findings, we propose~\name, a training-free frame selection framework for LMM that adapts its overall strategy based on the specific query type. The framework first employs an LLM to automatically classify a given query as either global or localized. For global queries, standard uniform sampling is employed. For localized queries, a highly targeted multi-stage pipeline is initiated. This pipeline begins with our proposed \textit{Content-Adaptive Frame Selection}, a method that leverages pairwise frame similarity based on DINO features~\cite{oquab2024dinov2learningrobustvisual} to select a set of semantically representative~\keyframes. Subsequently, the LMM itself is then utilized to score these~\keyframes, assigning a relevance reward based on their estimated utility in answering the query. Guided by this reward distribution, a video refinement process identifies and merges the most visually relevant video segments into a more condensed representation. Finally, this refined video is uniformly sampled to get the input frames for the LMM, ensuring that final inference is concentrated on the most pertinent temporal segments.

Our main contributions are summarized as follows:

\begin{itemize}[leftmargin=*]
    \item We identify a query typology (global vs. localized) and demonstrate that the efficacy of frame selection strategies is highly contingent on this classification.
    \item We propose~\name, a training-free frame selection framework that adapts to query type by employing uniform sampling for global queries and a specialized pipeline to extract query-relevant frames for localized queries.

    \item Experiments on three long-form video understanding benchmarks show that~\name~consistently outperforms existing baselines and robustly improves LMM's performance, even when scaling input frame count to 256.
\end{itemize}

%% file: sections/related_work.tex
\section{Related Work}
\label{sec:survey}

\subsection{Video-based Large Multimodal Models}

The rise of Transformer-based large language models (LLMs) has revolutionized natural language processing, with major advances stemming from increased model scale and larger pre-training datasets~\cite{dubeyLlama3Herd2024, openaiGPT4TechnicalReport2024a, brown2020languagemodelsfewshotlearners, touvron2023llamaopenefficientfoundation, peng2023instructiontuninggpt4, chowdhery2022palmscalinglanguagemodeling,chung2022scalinginstructionfinetunedlanguagemodels, zhang2022optopenpretrainedtransformer}. Inspired by this success, researchers have begun adapting LLMs to process multiple modalities, particularly integrating visual elements like images and videos~\cite{zhuMiniGPT4EnhancingVisionLanguage2023, li2024llavaonevisioneasyvisualtask, liu2023visualinstructiontuning, liu2024kangaroopowerfulvideolanguagemodel}, leading to the development of LMMs. Through extensive training, these models learn rich, cross-modal representations that effectively connect visual and textual information. This evolution has led to significant improvements across a range of video understanding applications, including tasks such as video captioning~\cite{yangVid2SeqLargeScalePretraining2023, chenShareGPT4VideoImprovingVideo2024, wuDIBSEnhancingDense2024, chai2025auroracapefficientperformantvideo, yan2023videococavideotextmodelingzeroshot} and video question answering~\cite{kimImageGridCan2024, minMoReVQAExploringModular2024, maaz2023video, chengVideoLLaMA2Advancing2024, lin2024videollavalearningunitedvisual,zhong2022videoquestionansweringdatasets}. Ongoing research is also focusing on refining model architectures~\cite{shi2025slowfastarchitecturevideomultimodal, xu2024slowfastllavastrongtrainingfreebaseline, zohar2024apolloexplorationvideounderstanding, shi2025eagleexploringdesignspace} and optimizing training strategies~\cite{zohar2024apolloexplorationvideounderstanding, liu2024kangaroopowerfulvideolanguagemodel} to further boost the performance of these systems. Despite their success, LMMs still struggle in video understanding due to the high volume of video tokens and the limited context length~\cite{song2024moviechatdensetokensparse, yen2024longcontextlanguagemodelingparallel}, as well as the ``Needle-in-a-Haystack'' issue~\cite{zhao2025needlevideohaystackscalable, ye2025rethinkingtemporalsearchlongform, li2025videochatflashhierarchicalcompressionlongcontext}. These challenges highlight the need for efficient frame selection techniques that capture key visual content without overloading the model.

\subsection{Video Token Reduction for the VQA Task}

In VQA task, uniform frame sampling is a standard technique for video token reduction. However, this method overlooks the query-specific relevance of individual frames. To address this limitation, recent research has focused on adaptive token reduction mechanisms, which are broadly classified into two primary categories.

\paragraph{Token Compression.}
This strategy carefully consolidates information within or across video frames to create a significantly more compact yet informative representation, reducing the total number of tokens needed~\cite{jin2024chatuniviunifiedvisualrepresentation}. Various advanced techniques are employed to achieve this, such as using a memory bank~\cite{songMovieChatDenseToken2024}, reducing temporal redundancy~\cite{shenLongVUSpatiotemporalAdaptive2024, wang2025retakereducingtemporalknowledge}, and applying hierarchical compression~\cite{li2025videochatflashhierarchicalcompressionlongcontext}.
Despite their inherent efficiency, token compression techniques may often lead to excessive summarization, resulting in the loss of critical fine-grained visual details. Moreover, highly query-related segments may be either compressed or overly generalized, ultimately compromising the model's capacity to effectively respond to the given query.

\begin{figure}[t]
    \centering
    \includegraphics[width=0.97\linewidth]{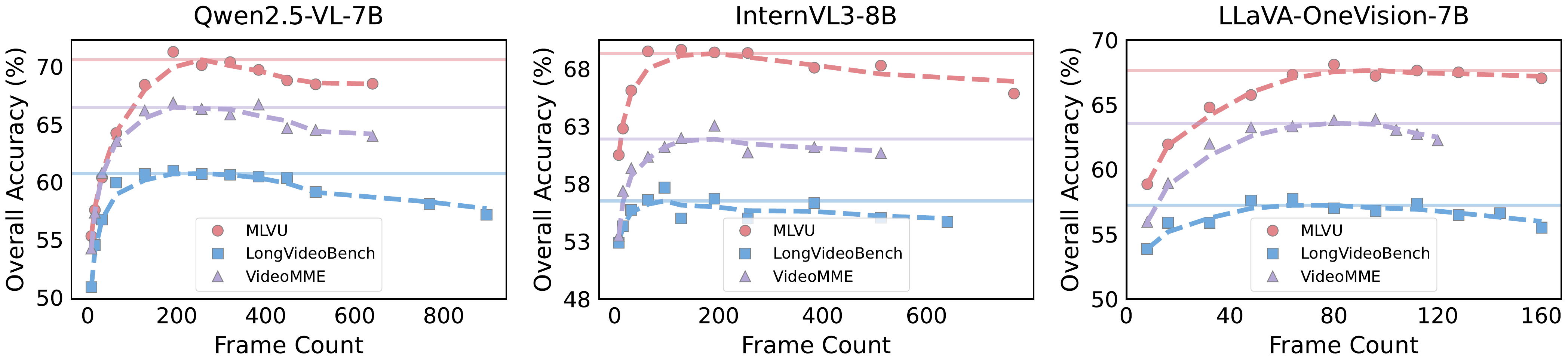}
    \caption{\textbf{Performance trends of various LMMs with respect to input frame counts.} We evaluate Qwen2.5-VL-7B~\cite{bai2025qwen25vltechnicalreport}, InternVL3-8B~\cite{chen2024internvl}, and LLaVA-OneVision-7B~\cite{li2024llavaonevisioneasyvisualtask} on MLVU~\cite{zhou2025mlvubenchmarkingmultitasklong}, LongVideoBench~\cite{wu2024longvideobenchbenchmarklongcontextinterleaved}, and VideoMME~\cite{fu2024videommefirstevercomprehensiveevaluation}. Results indicate that accuracy peaks at an optimal frame count and subsequently degrades, rather than improving monotonically.}
    \label{fig:cons}
    \vspace{-12pt}
\end{figure}

\paragraph{Query-Based Frame Selection.}

Compared with uniform sampling, recent methods employ more refined strategies to select query-relevant frames that typically involve three key steps: (1) uniformly sample candidate frames~\cite{liu2025boltboostlargevisionlanguage, tang2025adaptivekeyframesamplinglong, sun2025mdp3trainingfreeapproachlistwise, wang2024videoagentlongformvideounderstanding, wang2025videotreeadaptivetreebasedvideo, yang2025vcavideocuriousagent, fan2024videoagentmemoryaugmentedmultimodalagent} or video segments~\cite{ataallahGoldfishVisionLanguageUnderstanding2024a, ganz2024questionawarevisiontransformer}; (2) assess their relevance to the query using metrics like CLIPScore~\cite{hessel2022clipscorereferencefreeevaluationmetric, zhang2025qframequeryawareframeselection}, detector~\cite{ye2025trethinkingtemporalsearch} or learned models~\cite{yuFrameVoyagerLearningQuery2024}; (3) apply an algorithm to select the most relevant frames based on these scores. However, uniform sampling often balances poorly between information sparsity and computational load. Furthermore, relevance metrics like CLIPScore~\cite{hessel2022clipscorereferencefreeevaluationmetric} can be notoriously unreliable for complex reasoning, and the resulting temporally sparse frames may miss fine-grained details found in continuous clips. We address these limitations by: (1) using content-adaptive frame selection to identify frame candidates much more intelligently; (2) employing inherently more reliable LMMs for relevance assessment; and (3) retrieving and concatenating continuous clips corresponding to candidates before performing final frame selection, ensuring fine-grained information is  preserved.

%% file: sections/revisit.tex
\section{Revisiting Inference Mechanism of LMM in Video Understanding}
\label{sec:revisit}

Consider a video $V$ with $T$ frames, denoted as $\{f_i\}_{i=1}^T$, along with a query $Q$. In VQA task, the model receives the video $V$ and the query $Q$ as inputs and is tasked with generating a response $A$ that accurately addresses the query. In contemporary approaches, due to computational limitations and the language model's restricted context length $L$, only a subset of $N$ uniformly sampled frames, denoted as $\{f'_i\}_{i=1}^N$, is processed, where $N \ll T$. These selected frames are then combined with the query $Q$ and fed into the LMM, which autoregressively generates the answer $A$:
\begin{equation}
A = \text{LMM}([f'_1; f'_2; \ldots; f'_N; Q]).
\end{equation}
Obviously, a small subset of $N$ frames is often insufficient to capture the full content of a video, particularly in longer sequences. To address this, recent studies~\cite{chen2024longvilascalinglongcontextvisual, zhang2024longcontexttransferlanguage} have focused on extending model context lengths to allow more frames as input. However, this raises an important question: \textit{Does increasing the number of uniformly sampled input frames enhance performance on VQA task?}

\paragraph{More frames do not mean improved performance.}
To investigate this, we conducted an evaluation using three pretrained LMMs: Qwen2.5-VL-7B~\cite{bai2025qwen25vltechnicalreport}, InternVL3-8B~\cite{zhu2025internvl3exploringadvancedtraining}, and LLaVA-OneVision-7B~\cite{li2024llavaonevisioneasyvisualtask}, across three long-form video understanding benchmarks: MLVU~\cite{zhou2025mlvubenchmarkingmultitasklong}, VideoMME~\cite{fu2024videommefirstevercomprehensiveevaluation}, and LongVideoBench~\cite{wu2024longvideobenchbenchmarklongcontextinterleaved}. We employed uniform frame sampling with varying frame counts to evaluate the impact of frame count on model performance. As illustrated in Figure~\ref{fig:cons}, a consistent pattern emerges across all models and benchmarks: performance initially improves with more input frames but declines beyond a certain point.

\begin{figure}[t]
    \centering
    \includegraphics[width=0.97\linewidth]{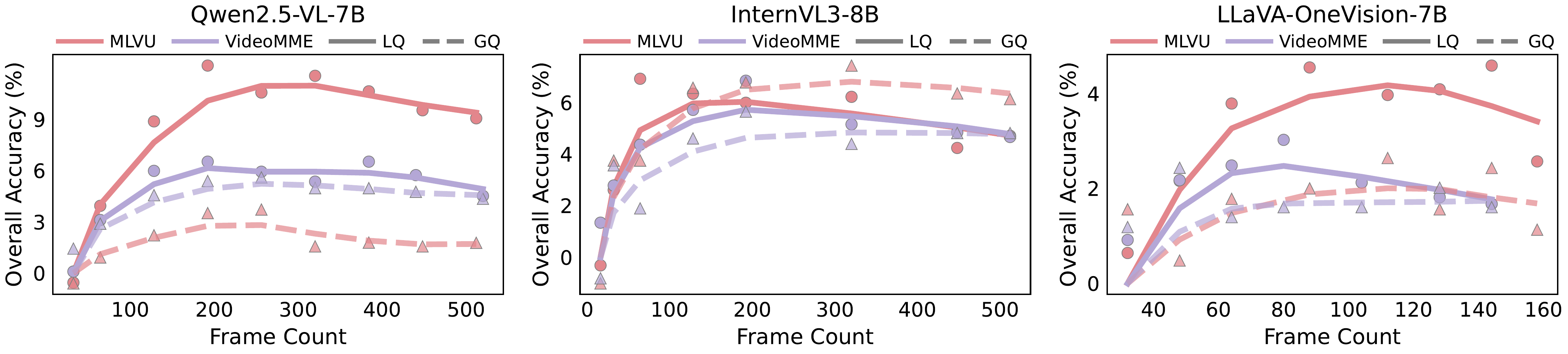}
  \caption{\textbf{Relative accuracy on localized and global queries.} Plotted as the deviation from the initial baseline, the results demonstrate that performance degradation at high frame counts is predominantly attributed to LQ, while GQ remains relatively stable.}
    \label{fig:gl}
    \vspace{-12pt}
\end{figure}

\paragraph{Query classification.}
To better understand the underlying causes of this performance degradation, we systematically examined the impact of different query types. Prior studies~\cite{liu2025boltboostlargevisionlanguage, tang2025adaptivekeyframesamplinglong, shen2024longvuspatiotemporaladaptivecompression, chen2024cgbenchcluegroundedquestionanswering, qu2025doesvisionlanguagemodellost} have already identified a class of queries that relate directly to specific, localized segments of a video, such as \textit{"What kind of bike is the man riding?"}, which we now classify as \textit{localized queries (LQ)}. However, these works frequently overlook another important category of queries requiring a comprehensive understanding of the entire video. We define such queries as \textit{global queries (GQ)}, with a typical example being \textit{"What title best summarizes this video?"}

\paragraph{Performance trends vary across query types.}
Following the definition, we manually categorize queries from MLVU~\cite{zhou2025mlvubenchmarkingmultitasklong} and VideoMME ~\cite{fu2024videommefirstevercomprehensiveevaluation}, and evaluate the same models on these two query types. As shown in Figure~\ref{fig:gl}, while performance on global queries remains relatively stable with increasing frame count, performance on localized queries drops significantly. We attribute this to global queries benefiting from holistic information, whereas excess frames introduce noise for localized tasks. These results highlight the necessity of pre-classifying query types to optimize efficiency; specifically, global queries can rely on standard uniform sampling, avoiding the computational overhead of key frame search techniques.

%% file: sections/method.tex
\section{Method: DIG}

\textbf{Overview.} In this section, we formally introduce~\name, a novel, training-free frame selection framework for LMMs that dynamically adapts to the query type.~\name~begins by classifying the given query as either localized or global (\S\ref{subsec:query_id}). For global queries, the final input frames are uniformly sampled across the entire video. In contrast, for localized queries, we first employ a content-adaptive frame selection method to extract highly representative frames (\S\ref{subsec:cafs}), which are then evaluated by the LMM through reward scoring to assess their relevance to the query (\S\ref{subsec:rw_as}). Then a refined video is carefully constructed through a search procedure guided by these rewards (\S\ref{subsec:search}) and final input frames are uniformly sampled from the refined video.

\subsection{Query Type Identification}
\label{subsec:query_id}
As established in Section~\ref{sec:revisit}, the performance trends vary across query types. Therefore, we first employ a LLM to classify a given query $Q$ as either global or localized (see Appendix~\ref{sec:prompt_de} for prompt details). For global queries, the LMM performs direct inference on uniformly sampled frames. Localized queries, in contrast, are addressed using the specialized approach detailed below.

\subsection{Content-Adaptive Frame Selection~(CAFS)}
\label{subsec:cafs}
To effectively address the localized query, it is essential to extract relevant frames from the video. However, exhaustive frame-wise analysis of long-form videos is computationally infeasible. This necessitates obtaining a compact yet informative subset of frames. Previous methods typically rely on static sampling (e.g., uniform or fixed-rate)~\cite{tang2025adaptivekeyframesamplinglong, liu2025boltboostlargevisionlanguage, ye2025trethinkingtemporalsearch, zhang2025qframequeryawareframeselection}. This static approach presents a dilemma: low-rate sampling may yield a sparse representation that misses critical events, while high-rate sampling produces a large and redundant frame set. To address this, we propose \textit{Content-Adaptive Frame Selection}, a method that adaptively selects representative frames, referred to as~\keyframes, based on high-level semantic content in the video such as objects and scenes.

\begin{figure}[t]
    \centering
    \includegraphics[width=0.97\linewidth]{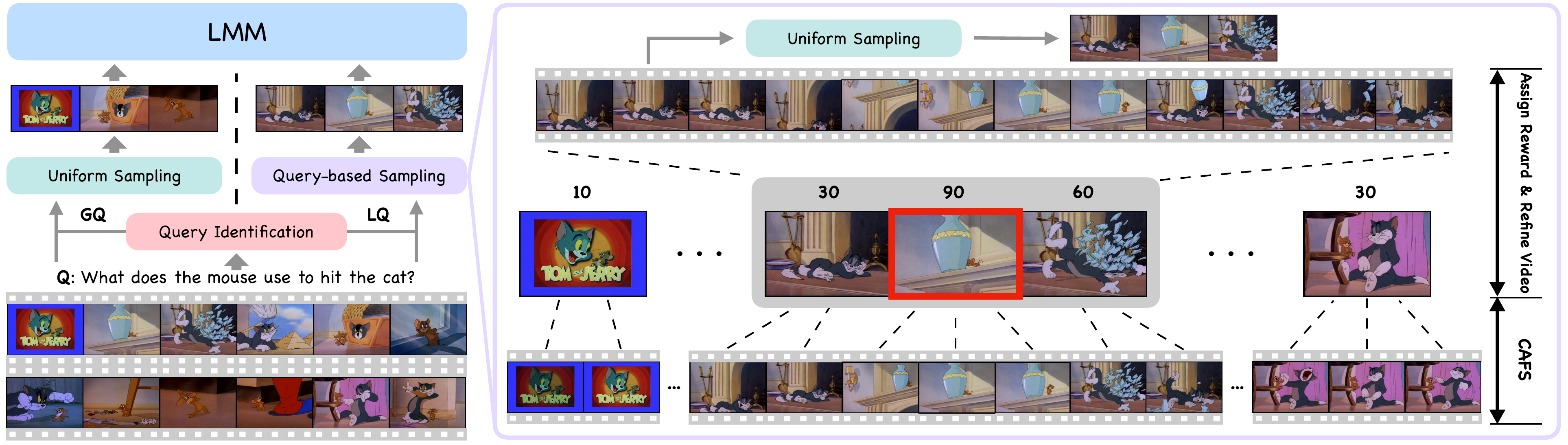}
    \caption{\textbf{Overview of~\name.} The LLM first classifies the query type. Global queries utilize uniform sampling across the entire video, while localized queries employ CAFS and reward assignment to construct a refined video prior to sampling. The selected frames are subsequently processed by the LMM for final inference.}
    \label{fig:pipeline_overview}
    \vspace{-12pt}
\end{figure}

\paragraph{Distance calculation.}

Given a 2-fps sampled video with $M$ frames $\{f_{I_i}\}_{i=1}^M$ with their corresponding frame indices $\{I_i\}_{i=1}^M$, we first utilize DINOv2~\cite{oquab2024dinov2learningrobustvisual} to extract robust visual features from each frame, which results in a sequence of feature vectors $\{V_{I_i}\}_{i=1}^M$. To accurately measure the dissimilarity between these consecutive frames, we compute the feature distance $d_i$ between $f_{I_i}$ and $f_{I_{i+1}}$ using the following formula:
\begin{equation}
    d_i = 1 - \text{sim}(V_{I_i}, V_{I_{i+1}}),
\end{equation}
where $\text{sim}(\cdot, \cdot)$ denotes cosine similarity. This yields a sequence of distances $\{d_i\}_{i=1}^{M-1}$.

\paragraph{R-Frame selection.}
Due to frequent scene transitions or camera cuts in long videos, the pairwise frame similarity often exhibits abrupt changes, resulting in numerous peaks in the distance sequence. Specifically, $d_i$ is identified as a peak if $d_{i-1} < d_i$ and $d_{i+1} < d_i$. To reduce noise effects, only peaks with prominence greater than $0.1$ are valid. This threshold has been found effective through empirical observation. We denote the indices of these valid peaks as $\{K_j\}_{j=1}^N \subset \{I_i\}_{i=1}^M$, where $N < M$. These peaks serve as segmentation points, dividing the video into distinct segments. Within each segment, the low pairwise distances between frames indicate visual consistency. Therefore, we select only one frame from each segment to capture its semantic content. For simplicity, we choose the midpoint frame of each segment, resulting in a set of~\keyframes~indexed by $\{I'_j\}_{j=1}^{N-1}=\left\{(K_j + K_{j+1})/2\right\}_{j=1}^{N-1}$. By aggregating~\keyframes, we obtain a compact representation that effectively summarizes the essential visual content of the entire video.

\subsection{Reward Assignment}
\label{subsec:rw_as}
To accurately identify the relevance of~\keyframes~to the given query $Q$, existing methods typically use either: (1) multimodal models like CLIPScore~\cite{radford2021learningtransferablevisualmodels,liu2025boltboostlargevisionlanguage, tang2025adaptivekeyframesamplinglong, wang2024videoagentlongformvideounderstanding}, or (2) object detection models to localize specific query-related entities in individual frames~\cite{ye2025rethinkingtemporalsearchlongform}. However, these traditional methods are often severely constrained by mere surface-level feature matching and reliance on fixed vocabularies, which fundamentally limits their ability to capture complex contextual reasoning and broader world knowledge. To address this, we directly leverage the LMM itself to assess frame relevance by assigning reward scores, with a simplified version of our prompt below.

\paragraph{Two-dimensional scoring.}
Since many queries, particularly those involving "why" or "how", cannot be fully addressed by a single frame, evaluating the relevance of individual frames independently may lead to incomplete or biased assessments. To mitigate this, we design the LMM to consider two complementary factors: (1) the direct relevance of the current frame to the query, and (2) whether the content of the current frame indicates that adjacent frames may contain supplementary information that contributes to a more comprehensive response.
\vspace{6pt}

\begin{tcolorbox}[title=Reward Model Prompt~(Simplified), fontupper=\small, fonttitle=\footnotesize, size=small, left=2.5mm,   
    right=2.5mm,  
    top=1mm,    
    bottom=1mm]
Frame: $<f_i>$; Query: $<Q>$; Please follow these steps to finish scoring:

1. Describe the sampled frame, focusing only on elements relevant to the question, if any.

2. Assign a relevance score between 0 and 100 based on:
(1) Direct usefulness of the frame for answering the query.
(2) Whether it suggests adjacent frames may contain relevant context.
\end{tcolorbox}

\subsection{Video Refinement}
\label{subsec:search}
Building upon the preceding steps, we have obtained the set of peak indices $\{K_j\}_{j=1}^N$, the~\keyframe~indices $\{I'_j\}_{j=1}^{N-1}$, and the reward values $\{R_j\}_{j=1}^{N-1}$ assigned to these~\keyframes. The next step is to select the most query-relevant~\keyframes~based on the reward values $\{R_j\}_{j=1}^{N-1}$.

\paragraph{Iterative reward-guided selection.}
In contrast to the commonly employed Top-K selection, which applies a fixed hyperparameter across varying scenarios, we introduce a parameter-free methodology. Given the initial rewards $\{R_j\}_{j=1}^{N-1}$, we iteratively refine this set until it stabilizes.

\begin{itemize}[leftmargin=*, nosep]
    \item \textit{Step 1.} Compute the mean of the current reward set: $\overline{R}$.
    \item \textit{Step 2.} Update each reward value by thresholding below the mean value:
    \begin{equation}
        R'_j = \max(R_j - \overline{R}, 0), \quad \forall j = 1,\ \dots,\ N-1.
    \end{equation}
    \item \textit{Step 3.} Finally, let $S$ be the resulting set of candidate indices $\{j \mid R'_j > 0\}$. Compare $S$ directly with the set of positive indices obtained from the previous iteration. If $S$ is strictly unchanged, terminate the entire iteration process. Otherwise, update the current reward set $\{R_j\} \gets \{R'_j\}$ and repeat from \textit{Step 1}.
\end{itemize}
Upon termination, the selected~\keyframes, denoted by $I_f$,  are formally defined as those~\keyframes~whose corresponding reward values in the final iteration are positive: $I_f = \{I'_j \mid R'_j > 0\}_{j=1}^{N-1}$. This criterion ensures that all~\keyframes~in the final selection set possess a reward larger than average.

\paragraph{Segment combination.}
Since~\keyframes~exhibit high feature similarity with their adjacent frames, it indicates an opportunity to incorporate fine-grained information beyond simply using them as input to the LMM. Specifically, for each selected~\keyframe~indexed by $I'_j$, we consider the video segment in the interval $[K_j, K_{j+1}]$ for richer temporal details. To capture more relevant context, we also consider adjacent~\keyframes~within a window of length $wlen$, specifically those with index range from $I'_{j - wlen}$ to $I'_{j + wlen}$. This results in the video segment spanning the index range $[K_{j - wlen}, K_{j + wlen + 1}]$. Then we combine the corresponding video segments of all selected~\keyframes~via union operation, resulting in a refined video containing query-relevant and fine-grained content. Finally, we uniformly sample frames from this refined video as input to the LMM.

%% file: sections/experiments.tex
\vspace{-6pt}
\section{Experiment}
\label{sec:experiment}
\subsection{Experiment Settings}

\paragraph{Datasets.}
We comprehensively evaluate our proposed approach on three benchmarks: MLVU~\cite{zhou2025mlvubenchmarkingmultitasklong}, LVB~\cite{wu2024longvideobenchbenchmarklongcontextinterleaved}, and VideoMME~\cite{fu2024videommefirstevercomprehensiveevaluation}, which contain complex videos ranging from several minutes to multiple hours, allowing us to assess long-form video understanding capabilities. For VideoMME~\cite{fu2024videommefirstevercomprehensiveevaluation}, we focus only on the medium and long splits. We don't use any subtitles, ensuring that evaluation is strictly based on pure visual understanding. Further benchmark details are provided in Appendix~\ref{sec:benchmark_details}.

\paragraph{Implementation details.}
The LMMs used are Qwen2.5-VL-7B~\cite{bai2025qwen25vltechnicalreport} and Qwen2.5-VL-32B~\cite{bai2025qwen25vltechnicalreport}. The LLM used for query identification is Qwen3-Next-80B-A3B~\cite{qwen3technicalreport}. Each input frame is represented using 56 tokens. The hyperparameter $wlen$ is set to 2. All experiments are conducted on 8 A100 GPUs within LMMs-Eval~\cite{zhang2024lmmsevalrealitycheckevaluation} framework. Additionally, we utilize vLLM backend~\cite{kwon2023efficient} to accelerate inference during the query identification and reward assignment stages. As baselines, we choose AKS~\cite{tang2025adaptivekeyframesamplinglong} and Q-Frame~\cite{zhang2025qframequeryawareframeselection}, and uniform sampling~(UNI). Detailed baseline configurations and extended experiments on Qwen3-VL-8B~\cite{bai2025qwen3vltechnicalreport} are available in Appendix~\ref{sec:detail_exp}.

\subsection{Main Results}

\paragraph{Comparison with existing methods.}
As shown in Table~\ref{tab:main}, compared with uniform sampling and competitive baselines including Q-Frame~\cite{zhang2025qframequeryawareframeselection} and AKS~\cite{tang2025adaptivekeyframesamplinglong},~\name~consistently improves performance on both Qwen2.5-VL-32B~\cite{bai2025qwen25vltechnicalreport} and Qwen2.5-VL-7B~\cite{bai2025qwen25vltechnicalreport} across input frame numbers from $8$ to $256$.
Notably, with $32$ frames,~\name~significantly boosts the accuracy of Qwen2.5-VL-7B~\cite{bai2025qwen25vltechnicalreport} by $7.68\%$ on MLVU~\cite{zhou2025mlvubenchmarkingmultitasklong} and $4.51\%$ on LongVideoBench~\cite{wu2024longvideobenchbenchmarklongcontextinterleaved} compared to uniform sampling.
This superiority extends to the more powerful Qwen2.5-VL-32B~\cite{bai2025qwen25vltechnicalreport}, where~\name~achieves better performance across almost all reported settings, effectively enhancing even a strong base model where other methods struggle to show consistent gains.

\begin{table}[t]
  \caption{\textbf{Performance comparison between different frame selection methods.} Base LMMs are Qwen2.5-VL-32B~\cite{bai2025qwen25vltechnicalreport} (left) and Qwen2.5-VL-7B~\cite{bai2025qwen25vltechnicalreport} (right). \textbf{Bold} indicates best performance, while \colorbox{red!10}{Red Box} denote results inferior to uniform sampling.}
  \label{tab:main}
  \centering

  \begin{minipage}{0.48\linewidth}
    \centering
    \resizebox{\linewidth}{!}{
      \begin{tabular}{l|c|ZZZZ} 
         \toprule
         \multirow{2}{*}{Method} & \multirow{2}{*}{\#Frames} & \multirow{2}{*}{MLVU} & \multirow{2}{*}{LVB} & \multicolumn{2}{c}{VideoMME} \\
         \cmidrule(lr){5-6}
         & & & & {Medium} & {Long}\\
         \midrule
         UNI & 8 & $55.93$ & $53.40$ & $53.89$ & $\textbf{51.56}$ \\
         Q-Frame~\cite{zhang2025qframequeryawareframeselection} & 8 & $56.03$ & $53.78$ & $54.03$ & \cellcolor{red!10}$49.63$ \\
         \textbf{DIG}~(\textit{Ours}) & 8 & $\textbf{61.55}$ & $\textbf{56.77}$ & $\textbf{54.12}$ & \cellcolor{red!10}$51.21$ \\
         \midrule
         UNI & 16 & $58.79$ & $54.67$ & $55.44$ & $\textbf{53.33}$ \\
         Q-Frame~\cite{zhang2025qframequeryawareframeselection} & 16 & \cellcolor{red!10}$57.73$ & $56.62$ & \cellcolor{red!10}$55.09$ & \cellcolor{red!10}$51.11$ \\
         \textbf{DIG}~(\textit{Ours}) & 16 & $\textbf{66.21}$ & $\textbf{58.86}$ & $\textbf{58.62}$ & \cellcolor{red!10}$52.18$ \\
         \midrule
         UNI & 32 & $61.91$ & $57.89$ & $57.89$ & $53.33$ \\
         AKS~\cite{tang2025adaptivekeyframesamplinglong} & 32 & $66.42$ & $59.31$ & $59.89$ & $56.00$ \\
         Q-Frame~\cite{zhang2025qframequeryawareframeselection} & 32 & \cellcolor{red!10}$60.95$ & \cellcolor{red!10}$57.37$ & $60.43$ & $55.90$ \\
         \textbf{DIG}~(\textit{Ours}) & 32 & $\textbf{70.69}$ & $\textbf{61.86}$ & $\textbf{60.87}$ & $\textbf{57.76}$ \\
         \midrule
         UNI & 64 & $66.24$ & $59.01$ & $64.33$ & $55.67$ \\
         AKS~\cite{tang2025adaptivekeyframesamplinglong} & 64 & $69.41$ & $61.41$ & $64.67$ & \textbf{58.44} \\
         Q-Frame~\cite{zhang2025qframequeryawareframeselection} & 64 & \cellcolor{red!10}$66.05$ & $59.61$ & \cellcolor{red!10}$62.80$ & $57.72$ \\
         \textbf{DIG}~(\textit{Ours}) & 64 & $\textbf{74.19}$ & $\textbf{63.65}$ & $\textbf{66.24}$ & $58.19$ \\
         \midrule
         UNI & 128 & $70.24$ & $61.78$ & $68.89$ & $59.67$ \\
         AKS~\cite{tang2025adaptivekeyframesamplinglong} & 128 & $72.77$ & $62.00$ & \cellcolor{red!10}$68.33$ & $61.44$ \\
         Q-Frame~\cite{zhang2025qframequeryawareframeselection} & 128 & \cellcolor{red!10}$70.10$ & \cellcolor{red!10}$60.06$ & \cellcolor{red!10}$68.21$ & \cellcolor{red!10}$59.28$ \\
         \textbf{DIG}~(\textit{Ours}) & 128 & $\textbf{75.20}$ & $\textbf{65.60}$ & $\textbf{69.00}$ & $\textbf{62.29}$ \\
         \midrule
         UNI & 192 & $71.76$ & $63.80$ & $69.56$ & $62.00$ \\
         AKS~\cite{tang2025adaptivekeyframesamplinglong} & 192 & $73.46$ & \cellcolor{red!10}$62.45$ & $69.89$ & \cellcolor{red!10}$61.00$ \\
         \textbf{DIG}~(\textit{Ours}) & 192 & $\textbf{76.66}$ & $\textbf{66.42}$ & $\textbf{70.11}$ & $\textbf{63.42}$ \\
         \bottomrule
      \end{tabular}
    }
  \end{minipage}
  \hfill
  \begin{minipage}{0.48\linewidth}
    \centering
    \resizebox{\linewidth}{!}{
      \begin{tabular}{l|c|ZZZZ} 
        \toprule
        \multirow{2}{*}{Method} & \multirow{2}{*}{\#Frames} & \multirow{2}{*}{MLVU} & \multirow{2}{*}{LVB} & \multicolumn{2}{c}{VideoMME} \\
        \cmidrule(lr){5-6}
        & & & & {Medium} & {Long}\\
        \midrule
        UNI & 8 & $53.64$ & $51.23$ & $51.36$ & $45.84$ \\
        Q-Frame~\cite{zhang2025qframequeryawareframeselection} & 8 & $54.42$ & $54.23$ & \cellcolor{red!10}$50.81$ & $\textbf{49.21}$ \\
        \textbf{DIG}~(\textit{Ours}) & 8 & $\textbf{58.64}$ & $\textbf{55.20}$ & $\textbf{54.23}$ & $46.88$ \\
        \midrule
        UNI & 16 & $56.43$ & $54.45$ & $55.94$ & $48.12$ \\
        Q-Frame~\cite{zhang2025qframequeryawareframeselection} & 16 & $56.81$ & $57.37$ & \cellcolor{red!10}$53.78$ & $49.02$ \\
        \textbf{DIG}~(\textit{Ours}) & 16 & $\textbf{63.98}$ & $\textbf{57.89}$ & $\textbf{56.81}$ & $\textbf{51.93}$ \\
        \midrule
        UNI & 32 & $59.52$ & $56.92$ & $59.08$ & $52.02$ \\
        AKS~\cite{tang2025adaptivekeyframesamplinglong} & 32 & $65.07$ & $59.31$ & $59.22$ & $53.11$ \\
        Q-Frame~\cite{zhang2025qframequeryawareframeselection} & 32 & $60.03$ & \cellcolor{red!10}$56.39$ & \cellcolor{red!10}$56.64$ & \cellcolor{red!10}$51.57$ \\
        \textbf{DIG}~(\textit{Ours}) & 32 & $\textbf{67.20}$ & $\textbf{60.43}$ & $\textbf{61.62}$ & $\textbf{53.24}$ \\
        \midrule
        UNI & 64 & $63.61$ & $58.94$ & $61.01$ & $51.27$ \\
        AKS~\cite{tang2025adaptivekeyframesamplinglong} & 64 & $66.59$ & $60.66$ & $\textbf{62.94}$ & $53.44$ \\
        Q-Frame~\cite{zhang2025qframequeryawareframeselection} & 64 & \cellcolor{red!10}$63.43$ & \cellcolor{red!10}$57.52$ & $61.32$ & $53.70$ \\
        \textbf{DIG}~(\textit{Ours}) & 64 & $\textbf{70.65}$ & $\textbf{61.41}$ & $62.61$ & $\textbf{55.30}$ \\
        \midrule
        UNI & 128 & $67.31$ & $61.86$ & $65.89$ & $54.84$ \\
        AKS~\cite{tang2025adaptivekeyframesamplinglong} & 128 & $68.68$ & \cellcolor{red!10}$60.36$ & \cellcolor{red!10}$65.67$ & $\textbf{55.93}$ \\
        Q-Frame~\cite{zhang2025qframequeryawareframeselection} & 128 & $68.03$ & \cellcolor{red!10}$59.76$ & $65.91$ & $54.81$ \\
        \textbf{DIG}~(\textit{Ours}) & 128 & $\textbf{71.40}$ & $\textbf{63.13}$ & $\textbf{66.78}$ & $55.69$ \\
        \midrule
        UNI & 192 & $69.03$ & $61.93$ & $67.01$ & $55.82$ \\
        AKS~\cite{tang2025adaptivekeyframesamplinglong} & 192 & $69.93$ & \cellcolor{red!10}$61.26$ & $\textbf{68.22}$ & \cellcolor{red!10}$54.41$ \\
        \textbf{DIG}~(\textit{Ours}) & 192 & $\textbf{72.32}$ & $\textbf{64.32}$ & $68.00$ & $\textbf{58.24}$ \\
        \midrule
        UNI & 256 & $69.15$ & $61.48$ & $66.31$ & $57.12$ \\
        AKS~\cite{tang2025adaptivekeyframesamplinglong} & 256 & $71.50$ & \cellcolor{red!10}$61.03$ & $67.56$ & \cellcolor{red!10}$55.11$ \\
        \textbf{DIG}~(\textit{Ours}) & 256 & $\textbf{72.46}$ & $\textbf{64.62}$ & $\textbf{67.66}$ & $\textbf{57.76}$ \\
        \bottomrule
      \end{tabular}
    }
  \end{minipage}
  \vspace{-6pt}
\end{table}

\begin{figure}[t]
    \centering
    \includegraphics[width=\linewidth]{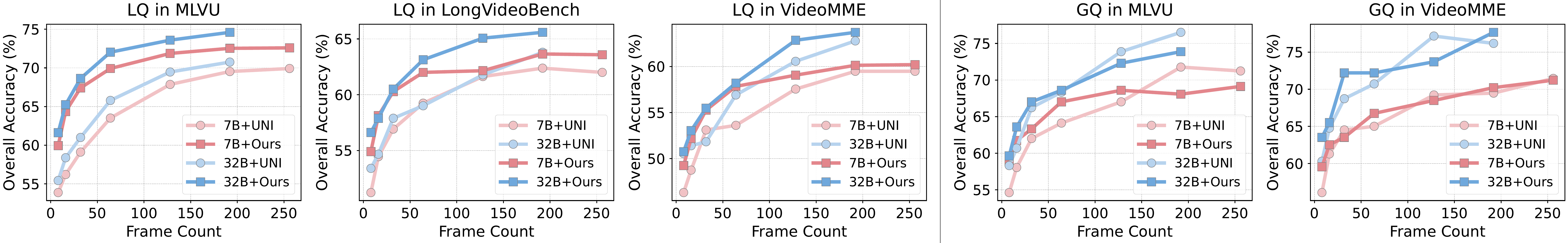}
    \caption{Comparison of our proposed frame selection pipeline (Sections~\ref{subsec:cafs}--\ref{subsec:search}) versus uniform sampling across different query types. The base LMMs are Qwen2.5-VL-7B~\cite{bai2025qwen25vltechnicalreport} and Qwen2.5-VL-32B~\cite{bai2025qwen25vltechnicalreport}.}
    \label{fig:abl_gl}
\end{figure}

\begin{figure}[t]
    \centering
    \includegraphics[width=0.95\linewidth]{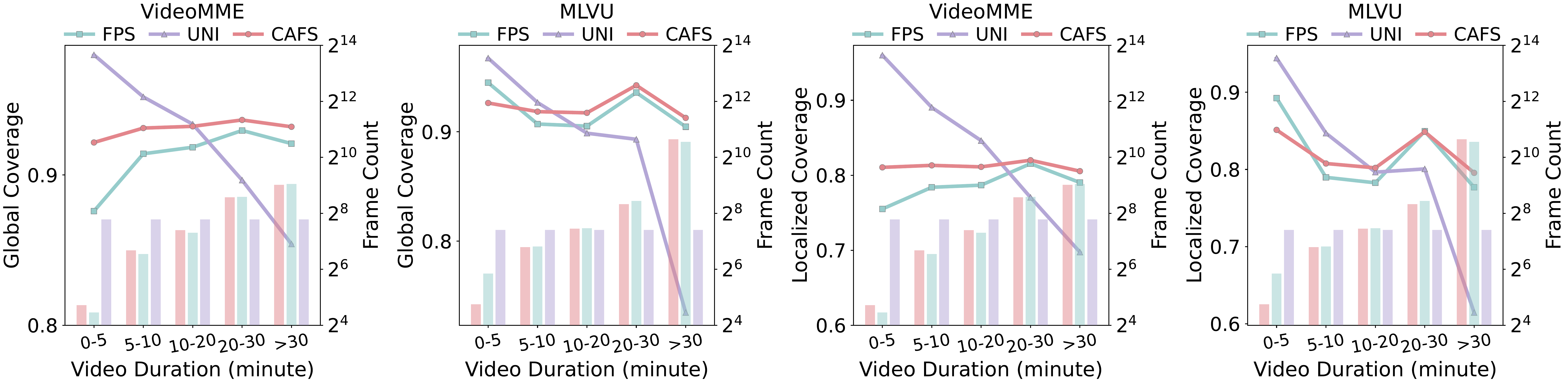}
    \caption{\textbf{GlC and LoC scores across varying video durations.} We compare three sampling strategies: FPS, UNI, and CAFS. In each sub-figure, the lines represent the score (left y-axis), while the bars indicate the number of sampled frames (right y-axis).}
    \label{fig:abl_cafs_gl}
    \vspace{-12pt}
\end{figure}

\paragraph{Scalability and performance consistency.}
In well-resourced environments, performance analysis at minimal frame counts (e.g., 8 or 16) offers limited practical insight, as applications typically seek to maximize frame utilization within given constraints. Therefore, unlike most previous works~\cite{tang2025adaptivekeyframesamplinglong, zhang2025qframequeryawareframeselection, liu2025boltboostlargevisionlanguage, yuFrameVoyagerLearningQuery2024, sun2025mdp3trainingfreeapproachlistwise} that validate performance in low-frame regimes ($<64$ frames), we conduct an evaluation that scales inputs to high frame densities (e.g., 256 frames). Under these conditions, as detailed in Table~\ref{tab:main}, AKS~\cite{tang2025adaptivekeyframesamplinglong} and Q-Frame~\cite{zhang2025qframequeryawareframeselection} can exhibit performance degradation relative to uniform sampling as frame counts increase. For instance, when utilizing the Qwen2.5-VL-7B~\cite{bai2025qwen25vltechnicalreport} with 128 input frames, both AKS~\cite{tang2025adaptivekeyframesamplinglong} and Q-Frame~\cite{zhang2025qframequeryawareframeselection} underperformed uniform sampling by 1--2\% on LongVideoBench~\cite{wu2024longvideobenchbenchmarklongcontextinterleaved}. In contrast, \name~demonstrates consistent performance gains over uniform sampling across all tested LMMs and most input frame configurations.

%% file: sections/discussion.tex
\section{Discussion and Analysis}
\label{sec:discussion}
To thoroughly evaluate the specific contributions of each individual module, in this section, we present a detailed analysis of~\name. Our evaluation is structured around the following key questions:

\begin{itemize}
    \item \textit{How does the choice of frame selection strategy impact performance on global versus localized queries?}~(\S\ref{subsec:abl_qi})
    \item \textit{How effective is the CAFS module at selecting representative frames, and what is its contribution to the overall performance of~\name?}~(\S\ref{subsec:abl_cafs})
    \item \textit{How does the LMM-based reward model compare against the CLIPScore~\cite{hessel2022clipscorereferencefreeevaluationmetric} in reward assignment?}~(\S\ref{subsec:abla_rw})
    \item \textit{What is the influence of the temporal window length (\textit{wlen}) on model performance?}~(\S\ref{subsec:wlen})
    \item \textit{What is the computational efficiency of~\name?}~(\S\ref{subsec:efficiency})
\end{itemize}

\subsection{Frame Selection Strategy vs. Query Type}
\label{subsec:abl_qi}
To examine the impact of different frame selection strategies on global versus localized queries, we leverage the query classifications established in Section~\ref{sec:revisit} and then compare the performance of uniform sampling against our proposed frame selection pipeline on each query type.

\paragraph{Efficacy of uniform sampling on GQ.}
As clearly illustrated in the right two charts of Figure~\ref{fig:abl_gl}, standard uniform sampling consistently achieves performance comparable to, or occasionally even superior to, our complex pipeline on GQs. This observation suggests that global queries generally necessitate comprehensive and temporally diverse information from the video content, which uniform sampling effectively provides.

\paragraph{Superiority of keyframe selection on LQ.} For LQs, our pipeline consistently outperforms uniform sampling, as shown in the left three charts of Figure~\ref{fig:abl_gl}. This result demonstrates our method's effectiveness in accurately identifying and extracting the specific video segments relevant to localized inquiries. These findings underscore the importance of a query-aware sampling strategy: identifying the query type is essential to determine whether to employ broad sampling for global context or targeted extraction for specific details.

\subsection{Analysis of CAFS Effectiveness}
\label{subsec:abl_cafs}

Let $f_j$ denote the frame indexed by $j$, and let $V_j$ represent its feature vector obtained via DINOv2~\cite{oquab2024dinov2learningrobustvisual}. We define the set of~\keyframes~as $\{f_{I_i}\}_{i=1}^N$ with indices $\{I_i\}_{i=1}^N$. To assess their effectiveness in capturing the high-level semantic content within a video, we introduce two quantitative metrics.

\paragraph{Localized Coverage~(LoC).}
This metric assesses the effectiveness with which each~\keyframe~captures its local temporal visual context. More specifically, for each~\keyframe~$f_{I_i}$, four neighboring frames are sampled uniformly from its surrounding temporal window. The LoC score is then computed as the average similarity between the~\keyframe~and its sampled neighbors across all~\keyframes.
\begin{equation}
\begin{aligned}
 \text{LoC} &= \frac{1}{4N} \sum_{i=1}^{N} \sum_{j=0}^{3} \text{sim}\left(V_{I_i}, V_{M_{i, j}}\right), \\
 \text{where } M_{i, j} &= I_i + \left(j-1.5\right) \cdot \left\lfloor (I_{i+1} - I_{i-1})/6 \right\rfloor
\end{aligned}
\end{equation}

\begin{table}[t]
  \centering
  \vspace{-6pt}
   \caption{Performance comparison with rewards from Qwen2.5-VL-32B~\cite{bai2025qwen25vltechnicalreport}, Qwen2.5-VL-7B~\cite{bai2025qwen25vltechnicalreport} and CLIPScore~\cite{hessel2022clipscorereferencefreeevaluationmetric} across various benchmarks. \textbf{Bold} indicates best performance. The base LMM used is Qwen2.5-VL-7B~\cite{bai2025qwen25vltechnicalreport}.}
   \label{tab:abl_rw}
    \resizebox{0.6\linewidth}{!}{
        \begin{tabular}{l|c|ZZZZZ} 
         \toprule
         \multirow{2}{*}{Method} & \multirow{2}{*}{\#Frames} & \multirow{2}{*}{MLVU} & \multirow{2}{*}{LVB} & \multicolumn{3}{c}{VideoMME} \\
         \cmidrule(lr){5-7}
         & & & & Short & {Medium} & {Long}\\
         \midrule
         CLIPScore~\cite{hessel2022clipscorereferencefreeevaluationmetric} & 8 & \cellcolor{lightgreen!15}57.4 & \cellcolor{lightgreen!15}52.6 & \cellcolor{lightgreen!15}62.3 & \cellcolor{lightgreen!15}51.1 & \cellcolor{lightgreen!15}\textbf{49.0} \\
         Qwen2.5-VL-7B~\cite{bai2025qwen25vltechnicalreport} & 8 & \cellcolor{lightblue!20}58.6 & \cellcolor{lightblue!20}55.2 & \cellcolor{lightblue!20}63.6 & \cellcolor{lightblue!20}\textbf{54.2} & \cellcolor{lightblue!20}46.9 \\
         Qwen2.5-VL-32B~\cite{bai2025qwen25vltechnicalreport} & 8 & \cellcolor{lightblue!20}\textbf{60.6} & \cellcolor{lightblue!20}\textbf{55.6} & \cellcolor{lightblue!20}\textbf{64.2} & \cellcolor{lightblue!20}52.6 & \cellcolor{lightblue!20}47.2 \\
         \midrule
         CLIPScore~\cite{hessel2022clipscorereferencefreeevaluationmetric} & 16 & \cellcolor{lightgreen!15}62.2 & \cellcolor{lightgreen!15}54.3 & \cellcolor{lightgreen!15}67.2 & \cellcolor{lightgreen!15}55.9 & \cellcolor{lightgreen!15}49.4 \\
         Qwen2.5-VL-7B~\cite{bai2025qwen25vltechnicalreport} & 16 & \cellcolor{lightblue!20}\textbf{64.0} & \cellcolor{lightblue!20}57.9 & \cellcolor{lightblue!20}67.8 & \cellcolor{lightblue!20}56.8 & \cellcolor{lightblue!20}\textbf{51.9} \\
         Qwen2.5-VL-32B~\cite{bai2025qwen25vltechnicalreport} & 16 & \cellcolor{lightblue!20}\textbf{64.0} & \cellcolor{lightblue!20}\textbf{59.2} & \cellcolor{lightblue!20}\textbf{68.1} & \cellcolor{lightblue!20}\textbf{57.6} & \cellcolor{lightblue!20}50.2 \\
         \midrule
         CLIPScore~\cite{hessel2022clipscorereferencefreeevaluationmetric} & 32 & \cellcolor{lightgreen!15}65.4 & \cellcolor{lightgreen!15}56.2 & \cellcolor{lightgreen!15}70.0 & \cellcolor{lightgreen!15}58.6 & \cellcolor{lightgreen!15}51.2 \\
         Qwen2.5-VL-7B~\cite{bai2025qwen25vltechnicalreport} & 32 & \cellcolor{lightblue!20}67.2 & \cellcolor{lightblue!20}60.4 & \cellcolor{lightblue!20}70.3 & \cellcolor{lightblue!20}\textbf{61.6} & \cellcolor{lightblue!20}\textbf{53.2} \\
         Qwen2.5-VL-32B~\cite{bai2025qwen25vltechnicalreport} & 32 & \cellcolor{lightblue!20}\textbf{67.9} & \cellcolor{lightblue!20}\textbf{60.6} & \cellcolor{lightblue!20}\textbf{72.6} & \cellcolor{lightblue!20}61.4 & \cellcolor{lightblue!20}53.1 \\
         \midrule
         CLIPScore~\cite{hessel2022clipscorereferencefreeevaluationmetric} & 64 & \cellcolor{lightgreen!15}67.2 & \cellcolor{lightgreen!15}59.6 & \cellcolor{lightgreen!15}72.7 & \cellcolor{lightgreen!15}62.4 & \cellcolor{lightgreen!15}54.7 \\
         Qwen2.5-VL-7B~\cite{bai2025qwen25vltechnicalreport} & 64 & \cellcolor{lightblue!20}70.7 & \cellcolor{lightblue!20}61.4 & \cellcolor{lightblue!20}73.3 & \cellcolor{lightblue!20}62.6 & \cellcolor{lightblue!20}\textbf{55.3} \\
         Qwen2.5-VL-32B~\cite{bai2025qwen25vltechnicalreport} & 64 & \cellcolor{lightblue!20}\textbf{71.0} & \cellcolor{lightblue!20}\textbf{63.4} & \cellcolor{lightblue!20}\textbf{74.4} & \cellcolor{lightblue!20}\textbf{64.8} & \cellcolor{lightblue!20}54.7 \\
         \midrule
         CLIPScore~\cite{hessel2022clipscorereferencefreeevaluationmetric} & 128 & \cellcolor{lightgreen!15}69.6 & \cellcolor{lightgreen!15}61.0 & \cellcolor{lightgreen!15}73.3 & \cellcolor{lightgreen!15}64.0 & \cellcolor{lightgreen!15}55.8 \\
         Qwen2.5-VL-7B~\cite{bai2025qwen25vltechnicalreport} & 128 & \cellcolor{lightblue!20}71.4 & \cellcolor{lightblue!20}63.1 & \cellcolor{lightblue!20}74.9 & \cellcolor{lightblue!20}66.8 & \cellcolor{lightblue!20}55.7 \\
         Qwen2.5-VL-32B~\cite{bai2025qwen25vltechnicalreport} & 128 & \cellcolor{lightblue!20}\textbf{72.6} & \cellcolor{lightblue!20}\textbf{65.2} & \cellcolor{lightblue!20}\textbf{75.4} & \cellcolor{lightblue!20}\textbf{69.2} & \cellcolor{lightblue!20}\textbf{57.1} \\
         \midrule
         CLIPScore~\cite{hessel2022clipscorereferencefreeevaluationmetric} & 192 & \cellcolor{lightgreen!15}71.0 & \cellcolor{lightgreen!15}62.5 & \cellcolor{lightgreen!15}74.6 & \cellcolor{lightgreen!15}63.9 & \cellcolor{lightgreen!15}54.8 \\
         Qwen2.5-VL-7B~\cite{bai2025qwen25vltechnicalreport} & 192 & \cellcolor{lightblue!20}72.3 & \cellcolor{lightblue!20}64.3 & \cellcolor{lightblue!20}75.9 & \cellcolor{lightblue!20}68.0 & \cellcolor{lightblue!20}\textbf{58.2} \\
         Qwen2.5-VL-32B~\cite{bai2025qwen25vltechnicalreport} & 192 & \cellcolor{lightblue!20}\textbf{73.9} & \cellcolor{lightblue!20}\textbf{65.4} & \cellcolor{lightblue!20}\textbf{76.2} & \cellcolor{lightblue!20}\textbf{69.2} & \cellcolor{lightblue!20}57.4 \\
         \midrule
         CLIPScore~\cite{hessel2022clipscorereferencefreeevaluationmetric} & 256 & \cellcolor{lightgreen!15}71.2 & \cellcolor{lightgreen!15}61.9 & \cellcolor{lightgreen!15}75.0 & \cellcolor{lightgreen!15}64.7 & \cellcolor{lightgreen!15}57.0 \\
         Qwen2.5-VL-7B~\cite{bai2025qwen25vltechnicalreport} & 256 & \cellcolor{lightblue!20}72.5 & \cellcolor{lightblue!20}\textbf{64.6} & \cellcolor{lightblue!20}76.3 & \cellcolor{lightblue!20}67.7 & \cellcolor{lightblue!20}57.8 \\
         Qwen2.5-VL-32B~\cite{bai2025qwen25vltechnicalreport} & 256 & \cellcolor{lightblue!20}\textbf{74.3} & \cellcolor{lightblue!20}64.5 & \cellcolor{lightblue!20}\textbf{76.8} & \cellcolor{lightblue!20}\textbf{68.9} & \cellcolor{lightblue!20}\textbf{59.1} \\
         \bottomrule
    \end{tabular}
    }
\end{table}

\begin{figure}[t]
    \centering
    \includegraphics[width=0.95\linewidth]{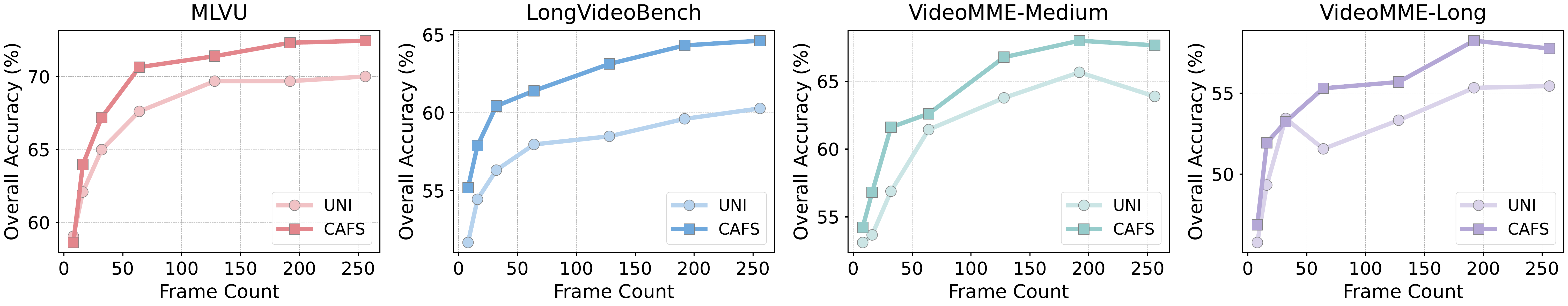}
    \caption{\textbf{Performance Comparison of CAFS and uniform sampling in~\name~pipeline.} The base LMM is Qwen2.5-VL-7B~\cite{bai2025qwen25vltechnicalreport}.}
    \label{fig:abl_cafs}
    \vspace{-6pt}
\end{figure}

\paragraph{Global Coverage~(GlC).}
This metric evaluates how well the~\keyframes~collectively represent the entire video content. Ideally, each frame in the video should be similar to at least one~\keyframe. To compute it, we randomly sample 200 frames from the video, denoted as $\{f_x\}_{x \in \mathcal{X}}$. For each frame $f_x$, we find the maximum similarity to any~\keyframe~and average these values across all sampled frames:
\begin{equation}
    \text{GlC} = \frac{1}{|\mathcal{X}|} \sum_{x \in \mathcal{X}} \max_{i \in [1, N]} \text{sim}(V_{I_i}, V_x)
\end{equation}

\paragraph{Baseline selection.}
We evaluate CAFS against two standard baselines: UNI (uniform frame sampling) and FPS (uniform frames-per-second sampling). The assessment is conducted on MLVU~\cite{zhou2025mlvubenchmarkingmultitasklong} and VideoMME~\cite{fu2024videommefirstevercomprehensiveevaluation}. To ensure a fair comparison, the average number of selected frames is kept consistent across all methods.

\paragraph{Analysis.}
As shown in Figure~\ref{fig:abl_cafs_gl}, the overall performance of standard uniform sampling declines with increasing video duration. This limitation arises from using a fixed number of frames across videos of varying lengths, which inevitably leads to significant redundancy in short videos and inadequate semantic coverage in long videos. Moreover, while regular fps sampling maintains stable performance, CAFS consistently outperforms it, particularly for videos over 10 minutes. This indicates that key semantic information in videos does not grow linearly with length, and that CAFS is more effective at selecting informative frames.

\paragraph{Comparison with uniform sampling in~\name.}
We compare CAFS with uniform sampling within the~\name~by replacing CAFS-extracted~\keyframes~with standard uniformly sampled ones. As experimentally shown in Figure~\ref{fig:abl_cafs}, CAFS robustly outperforms uniform sampling across all benchmarks. In addition, the observed performance gap widens with more input frames, further highlighting the fundamental limitation of uniform sampling: for long videos it cannot sample sufficient frames to adequately cover information for the subsequent reasoning process, while CAFS can effectively adapt to videos of any length and ensures significantly better coverage.

\subsection{Reward Assignment: LMM vs. CLIPScore}
\label{subsec:abla_rw}
We evaluate the reward assignment mechanism employed by the LMMs in~\name~by comparing it to a common alternative: computing frame-query similarity using CLIP~\cite{radford2021learningtransferablevisualmodels}. Specifically, we substitute all reward values originally assigned by the LMM with corresponding CLIPScore~\cite{hessel2022clipscorereferencefreeevaluationmetric}.

\paragraph{LMMs exhibit superior capability as reward assigners.}
As illustrated in Table~\ref{tab:abl_rw}, the rewards generated by LMMs (Qwen2.5-VL-7B/32B~\cite{bai2025qwen25vltechnicalreport}) demonstrate superior performance across the benchmarks in most cases, particularly as the number of frames increases. This underscores the LMM's capacity to deliver more precise and semantically rich reward signals through its advanced reasoning abilities and broad world knowledge. In contrast, CLIPScore~\cite{hessel2022clipscorereferencefreeevaluationmetric} depends on superficial feature matching and often fails to capture nuanced or visually complex query requirements.

\paragraph{Better LMMs yield superior rewards.}
The experimental results in Table~\ref{tab:abl_rw} also clearly indicate that employing the larger Qwen2.5-VL-32B~\cite{bai2025qwen25vltechnicalreport} as the reward assigner outperforms the smaller 7B variant, even on a short-video benchmark like VideoMME-short~\cite{fu2024videommefirstevercomprehensiveevaluation}. This confirms that more advanced LMMs provide considerably more precise reward signals, thereby facilitating more accurate identification of query-relevant frames. Furthermore, this directly highlights the inherent flexibility of our framework: we can effectively decouple the reward mechanism from the inference backbone. By leveraging a separate, reasoning-intensive Image-LMM for frame selection, we can significantly enhance the final performance of the primary Video-LMM.

\begin{figure}[t]
    \centering
    \includegraphics[width=0.98\linewidth]{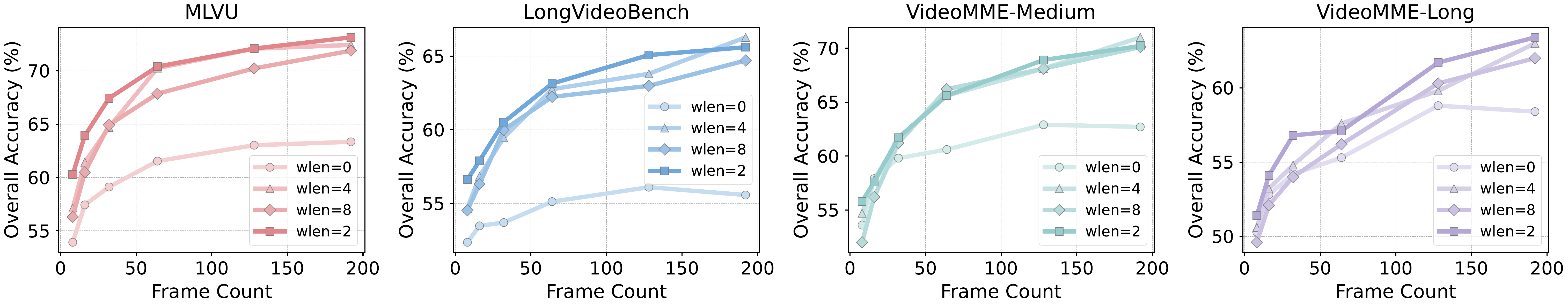}
    \caption{\textbf{Performance comparison of different window lengths~($wlen$) in~\name~pipeline.} The base LMM is Qwen2.5-VL-32B~\cite{bai2025qwen25vltechnicalreport}.}
    \label{fig:abla_alg}
    \vspace{-6pt}
\end{figure}

\subsection{Impact of Window Length}
\label{subsec:wlen}
To investigate how different values of $wlen$ affect performance, we conduct an evaluation using settings of $wlen \in \{0, 2, 4, 8\}$, while keeping all other settings identical.

\paragraph{Comparison with different window length.}
As shown in Figure~\ref{fig:abla_alg}, setting $wlen=0$ yields the lowest performance across all benchmarks. This deficit is particularly pronounced on LongVideoBench~\cite{wu2024longvideobenchbenchmarklongcontextinterleaved}, which necessitates reasoning over extended temporal contexts. This indicates that most queries cannot be effectively resolved within only a single scene, but instead require information from the surrounding temporal context. However, performance does not monotonically improve with $wlen$. When $wlen$ is set to a high value, such as $8$, performance degrades compared to $wlen=2$ and $4$. This proves that an excessively large window introduces irrelevant contextual information, creating noise that is detrimental to localized queries. Therefore, $wlen=2$ appears to strike the optimal balance, achieving the best results across the benchmarks.

\vspace{6pt}
\subsection{Efficiency of~\name}
\label{subsec:efficiency}

\begin{wrapfigure}{r}{0.45\linewidth}
\vspace{-14pt}
  \centering
    \includegraphics[width=\linewidth]{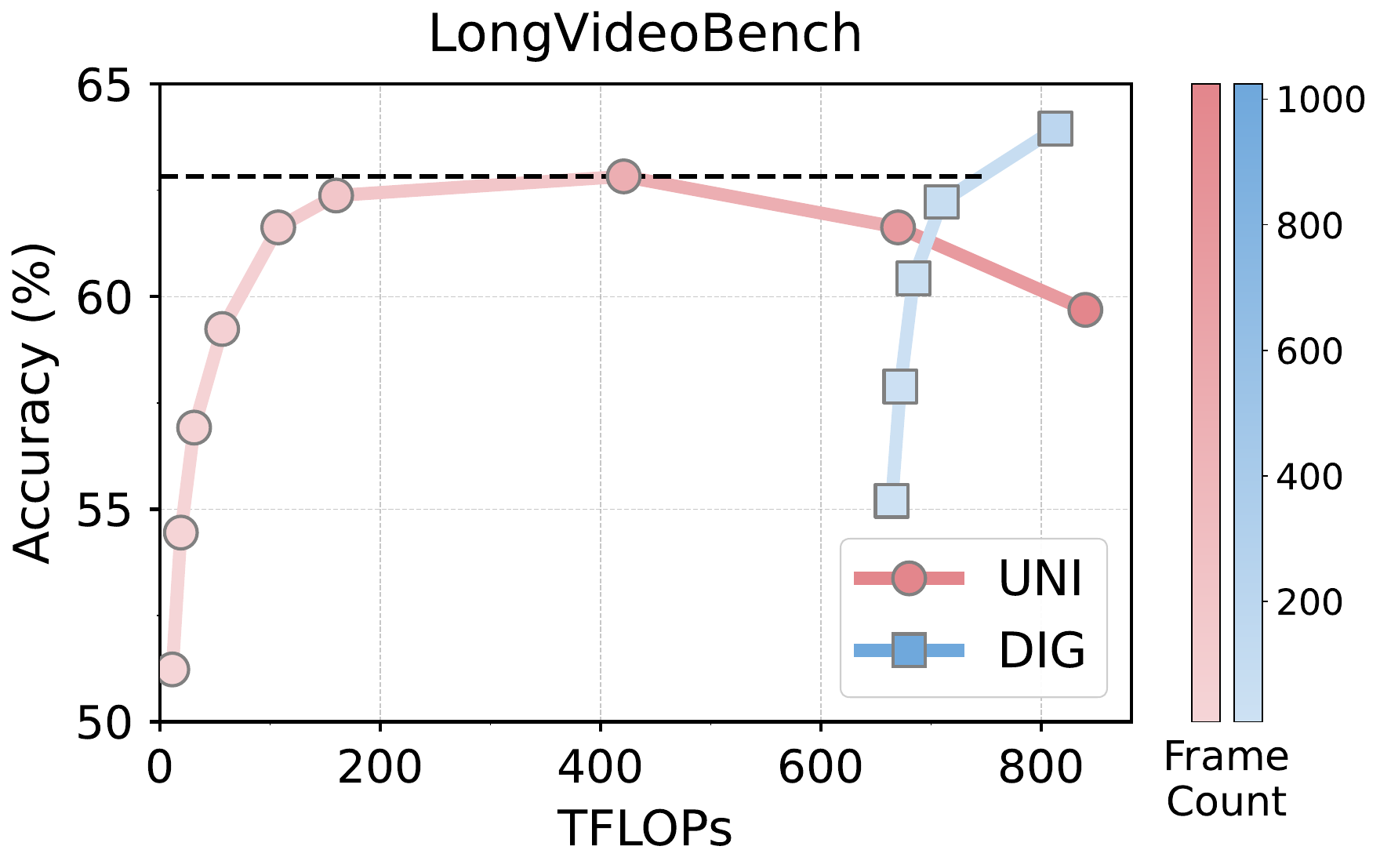}
    \caption{\textbf{Comparison between Accuracy and FLOPs.} The base LMM is Qwen2.5-VL-7B~\cite{bai2025qwen25vltechnicalreport}.}
    \label{fig:effciency}
\end{wrapfigure}

To evaluate computational cost, we measure and compare the FLOPs of our \name~pipeline against uniform sampling on LongVideoBench~\cite{wu2024longvideobenchbenchmarklongcontextinterleaved}. The reported FLOPs represent the average computation required per QA pair.

\paragraph{Performance-Efficiency analysis.}
As demonstrated in Figure~\ref{fig:effciency}, the uniform sampling approach exhibits a clear performance bottleneck. As the number of input frames scales, its accuracy saturates at a peak of $62.5\%$. Further increases in computation and frame count do not yield better performance. In contrast, \name~successfully overcomes this limitation. While operating at a higher computational budget ($\geq 680$ TFLOPs), \name~demonstrates positive performance scaling, surpassing the uniform sampling's peak accuracy once computation exceeds $720$ TFLOPs and continuing to improve thereafter.

%% file: sections/conclusion.tex
\section{Conclusion}
In this work, we find that optimal frame selection in video understanding depends on the query type (global vs. localized). Based on this, we propose~\name, a training-free framework that adapts to this typology: it employs efficient uniform sampling for global queries while reserving a multi-stage pipeline for localized queries where targeted selection is essential. This dual approach ensures both high performance and efficiency. Extensive experiments across diverse long-form video benchmarks and LMMs validate that~\name~consistently outperforms baselines and robustly scales LMM performance for inputs from 8 to 256 frames.

%% file: sections/appendix.tex
\noindent We include additional dataset statistics, annotation protocols, prompt designs, and extended experimental analyses to support reproducibility and offer deeper insight into each component of~\name. The appendix is organized as follows:

\begin{itemize}[leftmargin=7.5mm]
\setlength{\itemsep}{2pt}
\item Section~\ref{sec:benchmark_details} provides detailed statistics and descriptions of the benchmarks used in our experiments.
\item Section~\ref{sec:query_cls} describes the manual annotation protocol for classifying queries across each benchmark.
\item Section~\ref{sec:prompt_de} presents instructions used for query identification, reward assignment, and direct inference.
\item Section~\ref{sec:detailed_abl_qi} reports additional analysis on the query identification module, including LLM classification accuracy and its alignment with human annotations.
\item Section~\ref{sec:detailed_abl_cafs} elaborates on the CAFS algorithm and provides statistical analysis of its output characteristics.
\item Section~\ref{sec:detail_exp} contains per-task performance breakdowns on each benchmark, extended experiments on Qwen3-VL-8B, and further discussion of the results.
\item Section~\ref{sec:time} presents a detailed runtime profiling of~\name~and analyzes the efficiency gains from the query identification module.

\end{itemize}
\input{supp/1_benchmark}
\input{supp/2_classification}
\input{supp/3_prompt}
\input{supp/4_qi}
\input{supp/5_cafs}

\input{supp/6_exp_res}
\input{supp/7_time}

\begin{figure}[t]
    \centering
    \begin{tcolorbox}[title=Query Identification Prompt, fontupper=\small, fonttitle=\footnotesize]
    You are a helpful assistant in a video-based question-answering process.
    
    \vspace{6pt}
    \textbf{Core Task \& Definitions}
    
    You will classify the given query into one of two categories:

    \begin{enumerate}
        \item \textbf{Global Query (isGlobal: true):} The query requires going through and understanding the entire video content.
        \item \textbf{Localized Query (isGlobal: false):} The query that can be fully answered by extracting and analyzing several specific segments within the video.
    \end{enumerate}
    \vspace{6pt}
    \textbf{Instructions for Analysis and Response}
    
    In your analysis, please follow this structured reasoning process to classify the query:
    
    \textbf{Step 1. Understand the Query:} First, read the query to understand its general meaning and core intent.
    
    \textbf{Step 2. Infer Video Style (Hypothetically):} Based on the query's phrasing, make a reasonable inference about the style of the video (e.g., is it a narrative film, an educational lesson, a documentary, etc.)?
    
    \textbf{Step 3. Identify Referents:} Analyze if the query has specific referents. A referent is an entity (person, object), action, event, or even a specific piece of information, depending on the type of video you inferred. For instance, in 'What does Professor Smith write about quantum physics?', the referent is 'Professor Smith' and 'quantum physics' since the video style is likely a lesson.
    
    \textbf{Step 4. Evaluate Referents in Context:} Based on the results from step 3 and the criteria below, determine whether the query is Global or Localized.
    
    \begin{enumerate}[(i)]
        \item \textbf{The query is Global} if it meets either condition: 
        \begin{enumerate}[1.]
            \item Lacks a specific referent. The examples include: Summary-based: "primary focus," "in summary," "what is the video about?"
            \item Has a referent, but answering still requires a holistic understanding from going through the entire video. The examples include: "what is the boy's overall role?"
        \end{enumerate}
        
        \item \textbf{The query is Localized} if it has specific referents, and the answer can be found by focusing on specific, related segments where it appears. Here are some examples:
        \begin{itemize}
            \item Entity-based: "the person in the red shirt," "the black dog," "Professor Smith," "the little girl."
            \item Action/Event-based: "what is [X] doing," "how does [X] build,"
            \item Temporal/Sequential: "at the beginning," "after the explosion,"
        \end{itemize}
    \end{enumerate}

    Please provide your answer in the following format:
    \texttt{\{"analysis\_step1": str, "analysis\_step2": str, "analysis\_step3": str, "analysis\_step4": str, "isGlobal": true/false\}}

    \vspace{6pt}
    \textbf{User Query}: \textless Question\textgreater
    
    \end{tcolorbox}
     \vspace{-6pt}
    \caption{\textbf{Query Identification Prompt.} The LLM is first provided with the task definition, followed by an application of the chain-of-thought~\cite{wei2023chainofthoughtpromptingelicitsreasoning} technique to arrive at a judgment.}
    
    \label{fig:qi_prompt}
\end{figure}

\begin{figure}[h]
    \centering
    \begin{tcolorbox}[title=Reward Assignment Prompt, fontupper=\small, fonttitle=\footnotesize]
    You are a reward model for a video-based question-answering system.

    \vspace{6pt}
    \textbf{Task}
    
    You will receive a question and a sampled video frame. Your task is to evaluate the relevance of this frame for answering the question. Please assign a reward score that indicates how useful or informative the provided frame is in the context of the given question.

    \vspace{6pt}
    \textbf{Instructions for Analysis and Response}
    
    In your analysis, please perform the following steps to finish your evaluation:
    
    \begin{enumerate}
        \item Describe the visual content of the sampled frame, focusing on elements relevant to the question, if such elements are present.
        
        \item Assign a relevance reward between 0 and 100 based on: (1)~The sampled frame's direct usefulness in answering the question (2)~Whether the frame suggests that adjacent frames might provide additional information that help answer the question more effectively.
    \end{enumerate}
    
    Please provide your answer in the following format:
    \texttt{\{"description": str, "reward": int\}}.

    \vspace{6pt}
    \textbf{User Input}
    
    Video Duration: \textless Duration\textgreater\ seconds;
    Sampled Frame Timestamp: \textless Timestamp\textgreater\ seconds;
    Question: \textless Question\textgreater
    
    \end{tcolorbox}
    \vspace{-6pt}
    \caption{\textbf{Reward Assignment Prompt.} The LMM is first presented with the task definition and associated metadata. Then, the chain-of-thought reasoning technique~\cite{wei2023chainofthoughtpromptingelicitsreasoning} is applied to assign the reward for the input frame.}
    \label{fig:rw_prompt}
\end{figure}

\begin{figure}[t]
    \centering
    \begin{tcolorbox}[title=Inference Prompt]
    Question: What is the video mainly about?
    
    A. Planes invented by the Wright Brothers.
    
    B. The structural difference between the planes created by
    Whitehead and planes created by the Wright Brothers.
    
    C. Who invented the first plane.
    
    D. How Whitehead and the Wright Brothers cooperated to
    invent the first motorized flight.
    
    Please select the best answer from the options provided and directly provide the letter representing your choice without giving any explanation.
        
    \end{tcolorbox}
    \caption{\textbf{Prompt Template Example.} Example of the prompt template used by LMMs to perform direct inference.}
    \vspace{-6pt}
    \label{fig:exp_prompt}
\end{figure}

%% file: supp/1_benchmark.tex
\section{Benchmark Details} 
\label{sec:benchmark_details} 
This section details the benchmarks used in our evaluation. A statistical overview of each dataset is provided in Table~\ref{tab:benchmarks}.

\paragraph{MLVU.} 
MLVU~\cite{zhou2025mlvubenchmarkingmultitasklong} is a multi-task benchmark for long video understanding, comprising 3,102 questions across 9 categories. The dataset is partitioned into a dev set (2,593 questions) and a test set (509 questions). Tasks are categorized into three primary types: 1) holistic analysis, 2) single-detail identification, and 3) multi-detail reasoning. For our evaluation, we utilize only multiple-choice questions from the dev set and exclude open-ended questions.

\paragraph{LongVideoBench.} 
LongVideoBench~\cite{wu2024longvideobenchbenchmarklongcontextinterleaved} is a question-answering benchmark featuring 3,763 web-collected videos and 6,678 human-annotated, multiple-choice questions spanning 17 fine-grained categories. The benchmark is designed to test referring reasoning by requiring models to retrieve and reason over detailed information. In our study, we utilize only the validation set of this benchmark.

\paragraph{VideoMME.} 
VideoMME~\cite{fu2024videommefirstevercomprehensiveevaluation} is a multi-modal benchmark covering 30 subdomains across 6 primary visual domains. It contains 900 videos, totaling approximately 254 hours, and 2,700 question-answer pairs. The dataset includes multiple modalities (e.g., video, subtitles, audio) and splits videos by duration (short, medium, long). To focus our evaluation on long-form video understanding, we use only the medium and long duration splits. Furthermore, we leverage only the video data and corresponding questions, excluding all other modalities like subtitles.

\begin{table}[h]
    \centering
    \caption{\textbf{Dataset Statistics.} Overview of the data statistics across LongVideoBench~\cite{wu2024longvideobenchbenchmarklongcontextinterleaved}, MLVU~\cite{zhou2025mlvubenchmarkingmultitasklong} and VideoMME~\cite{fu2024videommefirstevercomprehensiveevaluation}.}
    \label{tab:benchmarks}
    \resizebox{0.6\linewidth}{!}{
    \begin{tabular}{l|c|c}
        \toprule 
        Dataset & Avg. Duration (s) & \#QA Pairs\\ 
        \midrule
        MLVU~\cite{zhou2025mlvubenchmarkingmultitasklong} & $636.2$ & $2174$ \\ LongVideoBench-val~\cite{wu2024longvideobenchbenchmarklongcontextinterleaved} & $732.2$ & $1337$\\
        VideoMME-short~\cite{fu2024videommefirstevercomprehensiveevaluation} & $80.7$ & $900$\\
        VideoMME-medium~\cite{fu2024videommefirstevercomprehensiveevaluation} & $516.8$ & $900$\\
        VideoMME-long~\cite{fu2024videommefirstevercomprehensiveevaluation} & $2466.3$ & $900$\\
        \bottomrule
    \end{tabular}
    }
\end{table}

\newpage

%% file: supp/2_classification.tex
\section{Query Identification by Human Annotator}
\label{sec:query_cls}
In this section, we elaborate on the query identification process described in Section~\ref{sec:revisit}, detailing the methodology used to classify queries from each benchmark.

\paragraph{MLVU.}
The task structure of MLVU~\cite{zhou2025mlvubenchmarkingmultitasklong} maps directly to our proposed query definitions. Queries associated with its "holistic tasks", which necessitate a comprehensive understanding of the entire video's overarching narrative, themes or a summary of its content, are classified as global queries. Conversely, queries within its "single-detail" and "multi-detail" task categories, which inherently demand that the model focus on specific, discrete temporal segments or isolated events, are classified as localized queries. Applying this classification scheme, we identified $462$ global queries and $1708$ localized queries within MLVU~\cite{zhou2025mlvubenchmarkingmultitasklong}.

\paragraph{LongVideoBench.}
The design of LongVideoBench~\cite{wu2024longvideobenchbenchmarklongcontextinterleaved} is centered on "referring reasoning." This evaluation paradigm is explicitly designed to test a model's capacity to ground its reasoning in specific, fine-grained visual information. By their very nature, such queries require pinpointing information within distinct temporal or spatial segments rather than assessing the video as a whole. Consequently, all queries within this benchmark correspond directly to our definition of localized queries.

\paragraph{VideoMME.}
VideoMME~\cite{fu2024videommefirstevercomprehensiveevaluation} lacks an intrinsic task classification that aligns with our global-versus-localized classification. To address this gap, we implemented a rigorous manual annotation process. We established a standardized protocol wherein human annotators were provided with detailed instructions and precise criteria (as illustrated in Figure~\ref{fig:qi_prompt}) to distinguish between the two query types. To ensure the reliability of these labels and mitigate subjective bias, the final classification for each query was determined by a majority vote consensus. This annotation procedure resulted in the identification of $479$ global queries and $2,221$ localized queries.

%% file: supp/3_prompt.tex
\section{Prompt Design}
\label{sec:prompt_de}

Prompt engineering is a cornerstone of harnessing the sophisticated reasoning capabilities of LLMs and LMMs. For our~\name~framework, we designed a series of specialized prompts to guide the models through our multi-stage video question-answering pipeline. This section details the design and rationale for the three core prompts: (1) Query Identification, (2) Reward Assignment, and (3) Direct Inference.

\paragraph{Query identification.} 
The initial and most critical step in our framework is to determine the type of the user's query. This classification dictates the subsequent processing strategy. As illustrated in Figure~\ref{fig:qi_prompt}, the prompt leverages a Chain-of-Thought (CoT) strategy~\cite{wei2023chainofthoughtpromptingelicitsreasoning} to deconstruct the classification task into a series of explicit, verifiable reasoning steps. The model is instructed to first analyze the query's intent, then hypothesize the video's genre (e.g., narrative, instructional), identify specific referents (entities, actions, or concepts), and finally synthesize this information to classify the query as either global or localized. This structured approach ensures a robust and transparent classification.

\paragraph{Reward assignment.} 
To generate fine-grained feedback for optimizing our video refinement process, we utilize an LMM to assign relevance scores to sampled frames. The prompt, shown in Figure~\ref{fig:rw_prompt}, presents the LMM with the user's question, a specific video frame, and associated metadata (video duration and frame timestamp). The model is tasked with a two-part CoT process: first, to provide a qualitative description of the frame's content, focusing on elements pertinent to the query, and second, to assign a quantitative reward score from 0 to 100. The reward criteria are carefully defined to capture not only the frame's direct usefulness but also its contextual value, that is, whether the frame suggests that temporally adjacent segments contain the necessary information.

\paragraph{Direct inference.} 
For final evaluation, we use a direct inference prompt, exemplified in Figure~\ref{fig:exp_prompt}. This prompt is designed for a standard multiple-choice question-answering format. It presents the LMM with the question and a set of candidate options (A, B, C, D). Additionally, the prompt instructs the model to return only the letter corresponding to the best answer.

\newpage

%% file: supp/4_qi.tex
\section{More Details about Query Identification}
\label{sec:detailed_abl_qi}

In this section, we evaluate the capability of contemporary LLMs to distinguish between global and localized queries. We assess the alignment between LLM predictions and human annotations by computing classification accuracy across three benchmarks: MLVU~\cite{zhou2025mlvubenchmarkingmultitasklong}, LongVideoBench~\cite{wu2024longvideobenchbenchmarklongcontextinterleaved}, and VideoMME~\cite{fu2024videommefirstevercomprehensiveevaluation}. The ground truth labels for these query types are derived from human annotations, as detailed in Section~\ref{sec:query_cls}.

\paragraph{LLMs exhibit strong alignment with human annotation.}
As presented in Table~\ref{tab:abl_qi}, nearly all evaluated LLMs achieve an overall classification accuracy exceeding $80\%$. This indicates that off-the-shelf LLMs possess sufficiently robust reasoning capabilities to effectively differentiate between localized and global queries without extensive fine-tuning when given a proper prompt.

\paragraph{Localized queries are more readily identifiable.}
Table~\ref{tab:abl_qi} further reveals that accuracy on localized queries consistently surpasses that of global queries. While GQ accuracy is comparatively lower, this has a negligible impact on final model performance; it primarily incurs a minor computational overhead. This is because, as established previously, performance differences between query-aware frame selection and uniform sampling are minimal for global queries. In addition, the critical metric is LQ accuracy that may influence the final performance. On this metric, almost all LLMs achieve an accuracy greater than 90\%, ensuring the final performance is good. And to make a tradeoff between compute cost and final model performance, we choose to use Qwen3-Next-80B-A3B-Instruct~\cite{qwen3technicalreport} in our main experiments.

\begin{table}[h]
    \centering
    \small  
    \caption{Accuracy (\%) of different LLMs in identifying localized queries (LQ) and global queries (GQ) across multiple benchmarks.}
    \label{tab:abl_qi}
    \resizebox{0.95\linewidth}{!}{
    \begin{tabular}{l|ccc|ccc|ccc}
        \toprule 
        \multirow{2}{*}{LLM} & \multicolumn{3}{c|}{MLVU~\cite{zhou2025mlvubenchmarkingmultitasklong}} & \multicolumn{3}{c|}{LongVideoBench~\cite{wu2024longvideobenchbenchmarklongcontextinterleaved}} & \multicolumn{3}{c}{VideoMME~\cite{fu2024videommefirstevercomprehensiveevaluation}} \\
        \cmidrule(lr){2-4} \cmidrule(lr){5-7} \cmidrule(lr){8-10}
        & LQ & GQ & \cellcolor{gray!20}Overall & LQ & GQ  & \cellcolor{gray!20}Overall & LQ & GQ  & \cellcolor{gray!20}Overall\\
        \midrule
        Qwen3-Next-80B-A3B-Instruct~\cite{qwen3technicalreport} & $87.02$ & $38.26$ &  \cellcolor{gray!20}$78.52$ & $97.53$ & N/A &  \cellcolor{gray!20}$97.53$ & $89.13$ & $65.76$ & \cellcolor{gray!20}$83.90$\\
        Llama-3.1-8B-Instruct~\cite{grattafiori2024llama3herdmodels} & $93.65$ & $24.01$ & \cellcolor{gray!20}$81.50$ & $98.20$ & N/A & \cellcolor{gray!20}$98.20$ & $96.99$ & $34.24$ & \cellcolor{gray!20}$82.95$ \\
        GPT-OSS-20B~\cite{openai2025gptoss120bgptoss20bmodel} & $82.00$ & $74.93$ & \cellcolor{gray!20}$80.77$ & $93.04$ & N/A & \cellcolor{gray!20}$93.04$ & $89.20$ & $69.97$ & \cellcolor{gray!20}$84.90$ \\
        DeepSeek-R1-Distill-Qwen-32B~\cite{deepseekai2025deepseekr1incentivizingreasoningcapability} & $93.03$ & $26.38$ & \cellcolor{gray!20}$81.42$ & $99.18$ & N/A & \cellcolor{gray!20}$99.18$ & $97.21$ & $52.85$ & \cellcolor{gray!20}$87.28$ \\
        \bottomrule
    \end{tabular}
    }
    \vspace{-12pt}
\end{table}

%% file: supp/5_cafs.tex
\section{More Details about CAFS}
\label{sec:detailed_abl_cafs}
This section provides a detailed examination of the CAFS method. Section~\ref{supp:subsec:alg_cafs} formally specifies the algorithm of CAFS, while Section~\ref{supp:subsec:res_cafs} presents a statistical analysis of its output characteristics based on practical application.

\subsection{Detailed Algorithm of CAFS}
\label{supp:subsec:alg_cafs}
Algorithm~\ref{alg:cafs} details our CAFS method. The process is structured into three sequential stages, taking a frame-to-frame distance sequence $d = [d_1, \ldots, d_{M-1}]$ and their corresponding original frame indices $I = [I_1, \ldots, I_M]$ as input, to produce a final set of~\keyframe~indices, \texttt{r\_idx}.

\paragraph{Initial peak detection.}
First, we identify all potential content boundaries. It iterates through the distance sequence, identifying any point $d_i$ that is a local maximum, defined as being greater than its two immediate neighbors ($d_{i-1} < d_i< d_{i+1}$). The indices $i$ of all such local maxima are collected into an initial \texttt{peaks} set.

\paragraph{Topographic prominence filtering.}
Second, we prune the \texttt{peaks} set, retaining only the most significant content transitions. For each peak $j \in \texttt{peaks}$, it calculates its "prominence" by finding the lowest base levels to its left ($l_{\text{min}}$) and right ($r_{\text{min}}$). The prominence is then defined as the peak's height $d_j$ minus the higher of its two bases ($\texttt{prominence} = d_j - \max(l_{\text{min}}, r_{\text{min}})$). This metric quantifies how much a peak "stands out" from the surrounding distance signal. Only peaks whose \texttt{prominence} exceeds a threshold (e.g., 0.1) are added to the \texttt{filtered\_peaks} set, effectively discarding minor, localized fluctuations.

\paragraph{R-Frame selection.}
Finally, we generate the output by identifying frames that best represent the stable content \textit{between} these significant transitions. The algorithm iterates through consecutive pairs of prominent peaks ($p_1, p_2$) from the filtered set. For each pair, it calculates the temporal midpoint using their associated original frame indices from $I$: $\texttt{midpoint} = (I_{p_1} + I_{p_2}) / 2$. These midpoints, which correspond to the center of the most stable segments, are aggregated into the final \texttt{r\_idx} set.

\begin{algorithm}[t]

\SetKwComment{Comment}{// }{}
\SetCommentSty{texttt}

\KwIn{Distance sequence $d = [d_1, d_2, \ldots, d_{M-1}]$, Frame indices $I = [I_1, I_2, \ldots, I_M]$}
\KwOut{Selected r-frame indices $\mathcal{R}_{\text{idx}}$}
\caption{Content-Adaptive Frame Selection}
\label{alg:cafs}

$\mathcal{P} \gets \emptyset$\;
\For{$i = 2$ \KwTo $M - 2$}{
    \If{$d_{i-1} < d_i$ \textnormal{and} $d_i > d_{i+1}$}{
        $\mathcal{P} \gets \mathcal{P} \cup \{i\}$ \Comment*[r]{A peak is a point higher than its neighbors}
    }
}

$\mathcal{P}_{\text{valid}} \gets \emptyset$\;
\ForEach{$j \in \mathcal{P}$}{
    $l_{\text{min}} \gets d_j$\;
    $k \gets j-1$\;
    \While{$k \geq 1$ \textnormal{and} $d_k \leq d_j$}{
        $l_{\text{min}} \gets \min(l_{\text{min}}, d_k)$\;
        $k \gets k-1$\;
    }
    
    $r_{\text{min}} \gets d_j$\;
    $m \gets j+1$\;
    \While{$m \leq M-1$ \textnormal{and} $d_m \leq d_j$}{
        $r_{\text{min}} \gets \min(r_{\text{min}}, d_m)$\;
        $m \gets m+1$\;
    }
    
    $p_{\text{prom}} \gets d_j - \max(l_{\text{min}}, r_{\text{min}})$ \Comment*[r]{Calculate topographic prominence}
    \If{$p_{\text{prom}} > 0.1$}{
        $\mathcal{P}_{\text{valid}} \gets \mathcal{P}_{\text{valid}} \cup \{j\}$\;
    }
}

$\mathcal{R}_{\text{idx}} \gets \emptyset$\;
\For{$i = 1$ \KwTo $|\mathcal{P}_{\text{valid}}| - 1$}{
    $p_1 \gets \mathcal{P}_{\text{valid}}[i]$\;
    $p_2 \gets \mathcal{P}_{\text{valid}}[i+1]$\;
    $m \gets (I_{p_1} + I_{p_2}) / 2$ \Comment*[r]{Midpoints between consecutive prominent peaks}
    $\mathcal{R}_{\text{idx}} \gets \mathcal{R}_{\text{idx}} \cup \{m\}$\;
}

\Return{$\mathcal{R}_{\text{idx}}$}\;
\end{algorithm}
\vspace{-6pt}

\subsection{More Results of CAFS}
\label{supp:subsec:res_cafs}
To further analyze the performance of CAFS on specific examples, we conduct an evaluation about the relationship between the number of~\keyframes~and video duration.

\begin{figure}[t]
    \centering
    \includegraphics[width=0.97\linewidth]{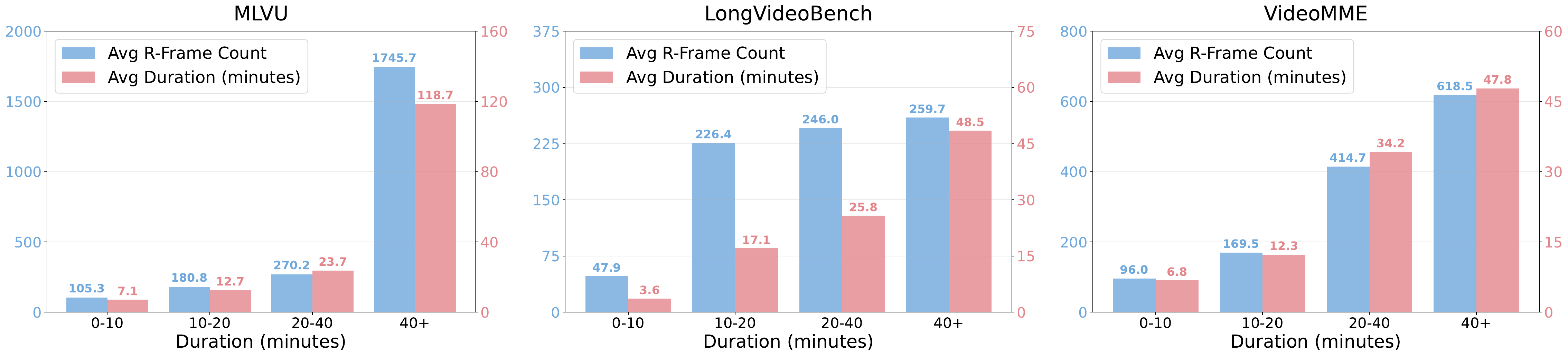} 
    \vspace{-6pt}
    \caption{Correlation between video duration and the number of~\keyframes~selected by the CAFS method across different benchmarks.}
    
    \label{fig:cafs_stats}
\end{figure}

\paragraph{Non-Linear information scaling in videos.}
Figure~\ref{fig:cafs_stats} reveals that the~\keyframe~count does not scale linearly with video duration. This non-linearity is prominent in LongVideoBench~\cite{wu2024longvideobenchbenchmarklongcontextinterleaved}: videos in the $0-10$ minute bracket average $47.9$~\keyframes, whereas those in the $10-20$ minute bracket average $226.4$. This finding exposes a fundamental limitation of fixed-rate sampling strategies (e.g., $N$ frames/video or $M$ frames/sec). Such approaches implicitly assume a uniform information distribution, leading to a suboptimal trade-off: sparse sampling risks information loss, while dense sampling incurs high temporal redundancy. CAFS bypasses this limitation by dynamically adapting its selection to the video's content density.

\paragraph{High context compression efficiency.} 
CAFS effectively condenses prolonged video-level context into a sparse, salient set of~\keyframes. For instance, on MLVU~\cite{zhou2025mlvubenchmarkingmultitasklong}, videos in the $10-20$ minute bracket ($12.7$ min avg.) are reduced to just $180.8$~\keyframes~on average. This represents a sparse sampling interval of approximately one~\keyframe~every $4.22$ seconds, demonstrating CAFS's capability to efficiently distill essential information from extended video sequences.

%% file: supp/6_exp_res.tex
\section{More Details about Experiment}
\label{sec:detail_exp}

\subsection{Detailed Experiment Settings}

\paragraph{Baseline setup.}
For AKS~\cite{tang2025adaptivekeyframesamplinglong}, we adhered to the default configuration: candidate frames were sampled at 1 fps, and frame-question similarity was computed via BLIP~\cite{li2022blipbootstrappinglanguageimagepretraining}. Based on the algorithm's selection logic, we evaluated frame budgets of $\{32, 64, 128, 192, 256\}$. We excluded budgets of $8$ and $16$ as the algorithm occasionally yielded null returns at these low settings. For Q-Frame~\cite{zhang2025qframequeryawareframeselection}, we employed the default "fixed frame count" strategy. Since this method limits the initial candidate pool to 128 frames, our evaluation was restricted to budgets of $\{8, 16, 32, 64, 128\}$.

\begin{table}[t]
  \caption{\textbf{Performance comparison between different frame selection methods.} Base LMM is Qwen3-VL-8B~\cite{bai2025qwen3vltechnicalreport}. \textbf{Bold} indicates best performance, while \colorbox{red!10}{Red Box} denote results inferior to uniform sampling.}
  \label{tab:qwen3vl}
  \centering
    \resizebox{0.5\linewidth}{!}{
      \begin{tabular}{l|c|ZZZZ}
         \toprule
         \multirow{2}{*}{Method} & \multirow{2}{*}{\#Frames} & \multirow{2}{*}{MLVU} & \multirow{2}{*}{LVB} & \multicolumn{2}{c}{VideoMME} \\
         \cmidrule(lr){5-6}
         & & & & {Medium} & {Long}\\
         \midrule
         UNI & 8 & $53.4$ & $50.3$ & $48.4$ & $49.2$ \\
         \textbf{DIG}~(\textit{Ours}) & 8 & $\textbf{58.2}$ & $\textbf{54.9}$ & $\textbf{53.1}$ & $49.8$ \\
         \midrule
         UNI & 16 & $53.9$ & $50.9$ & $51.3$ & $48.0$ \\
        \textbf{DIG}~(\textit{Ours}) & 16 & $\textbf{58.9}$ & $\textbf{53.9}$ & $\textbf{52.9}$ & $\textbf{49.9}$ \\
         \midrule
         UNI & 32 & $53.7$ & $51.0$ & $49.4$ & $48.6$ \\
         AKS~\cite{tang2025adaptivekeyframesamplinglong} & 32 & $57.3$ & $\textbf{54.4}$ & $52.3$ & $\textbf{50.1}$ \\
         \textbf{DIG}~(\textit{Ours}) & 32 & $\textbf{58.7}$ & $53.2$ & $\textbf{53.9}$ & $49.4$ \\
         \midrule
         UNI & 64 & $54.7$ & $51.2$ & $49.9$ & $48.9$ \\
         AKS~\cite{tang2025adaptivekeyframesamplinglong} & 64 & $56.3$ & $52.7$ & $50.9$ & $\textbf{51.6}$ \\
         \textbf{DIG}~(\textit{Ours}) & 64 & $\textbf{59.6}$ & $\textbf{54.8}$ & $\textbf{54.7}$ & $49.0$ \\
         \midrule
         UNI & 128 & $57.2$ & $54.4$ & $55.1$ & $\textbf{51.3}$ \\
         AKS~\cite{tang2025adaptivekeyframesamplinglong} & 128 & $58.9$ & \cellcolor{red!10}$53.0$ & \cellcolor{red!10}$54.4$ & \cellcolor{red!10}$51.1$ \\
         \textbf{DIG}~(\textit{Ours}) & 128 & $\textbf{64.4}$ & $\textbf{58.3}$ & $\textbf{58.4}$ & \cellcolor{red!10}$51.1$ \\
         \midrule
         UNI & 192 & $58.9$ & $57.1$ & $57.6$ & $51.6$ \\
         AKS~\cite{tang2025adaptivekeyframesamplinglong} & 192 & $61.1$ & \cellcolor{red!10}$54.5$ & $\textbf{61.0}$ & $\textbf{53.8}$ \\
         \textbf{DIG}~(\textit{Ours}) & 192 & $\textbf{66.8}$ & $\textbf{60.4}$ & $58.2$ & \cellcolor{red!10}$50.7$ \\
         \midrule
         UNI & 256 & $60.4$ & $57.6$ & $57.8$ & $53.4$ \\
         AKS~\cite{tang2025adaptivekeyframesamplinglong} & 256 & $63.8$ & \cellcolor{red!10}$55.6$ & $59.2$ & \cellcolor{red!10}$53.2$ \\
         \textbf{DIG}~(\textit{Ours}) & 256 & $\textbf{69.0}$ & $\textbf{61.2}$ & $\textbf{61.6}$ & $\textbf{53.8}$ \\
         \midrule
         UNI & 512 & $65.4$ & $60.2$ & $61.4$ & $55.0$ \\
         AKS~\cite{tang2025adaptivekeyframesamplinglong} & 512 & $65.5$ & \cellcolor{red!10}$57.6$ & \cellcolor{red!10}$60.7$ & $56.0$ \\
         \textbf{DIG}~(\textit{Ours}) & 512 & $\textbf{71.7}$ & $\textbf{63.8}$ & $\textbf{65.6}$ & $\textbf{56.4}$ \\
         \midrule
         UNI & 768 & $67.5$ & $60.9$ & $64.3$ & $56.6$ \\
         AKS~\cite{tang2025adaptivekeyframesamplinglong} & 768 & \cellcolor{red!10}$65.3$ & \cellcolor{red!10}$58.3$ & \cellcolor{red!10}$62.3$ & $57.3$ \\
         \textbf{DIG}~(\textit{Ours}) & 768 & $\textbf{72.2}$ & $\textbf{64.6}$ & $\textbf{67.8}$ & $\textbf{59.0}$ \\
         \bottomrule
    \end{tabular}
    }
\end{table}

\begin{table}[t] 
   \caption{\textbf{Performance comparison between different frame selection methods on MLVU.} Base LMMs are Qwen2.5-VL-32B~\cite{bai2025qwen25vltechnicalreport} (left) and Qwen2.5-VL-7B~\cite{bai2025qwen25vltechnicalreport} (right). \textbf{Bold} indicates best performance. The tasks of MLVU~\cite{zhou2025mlvubenchmarkingmultitasklong} are PlotQA~(PQA), NeedleQA~(NQA), Action Count~(AC), Action Order~(AO), Ego Reasoning~(ER), Anomaly Recognition~(AR), Topic Reasoning~(TR).} 
  \label{tab:exp_mlvu} 
  \centering 
  \small
  \begin{minipage}{0.49\linewidth} 
    \resizebox{\linewidth}{!}{
    \begin{tabular}{l|c|ccccccc}
        \toprule
        \multirow{2}{*}{Method} & \multirow{2}{*}{\#Frames} &  \multicolumn{7}{c}{MLVU~\cite{zhou2025mlvubenchmarkingmultitasklong}} \\ 
        \cmidrule(lr){3-9}
        & & PQA & NQA & AC & AO & ER & AR & TR\\
        \midrule
        UNI & 8 & $55.8$ & $58.6$ & $18.5$ & $51.4$ & $50.6$ & $66.5$ & $\textbf{85.6}$ \\
        Q-Frame~\cite{zhang2025qframequeryawareframeselection} & 8 & $51.4$ & $63.9$ & $18.4$ & $\textbf{60.2}$ & $50.3$ & $\textbf{70.5}$ & $76.8$ \\
        \rowcolor{gray!20} \textbf{DIG}~(\textit{Ours}) & \cellcolor{gray!20}8 & \cellcolor{gray!20}$\textbf{62.3}$ & \cellcolor{gray!20}$\textbf{73.0}$ & \cellcolor{gray!20}$\textbf{27.2}$ & \cellcolor{gray!20}$58.3$ & \cellcolor{gray!20}$\textbf{56.2}$ & \cellcolor{gray!20}$66.0$ & \cellcolor{gray!20}$84.0$ \\
        \midrule
        UNI & 16 & $59.4$ & $63.9$ & $18.0$ & $54.8$ & $52.8$ & $69.5$ & $86.3$ \\
        Q-Frame~\cite{zhang2025qframequeryawareframeselection} & 16 & $56.4$ & $64.8$ & $19.9$ & $59.5$ & $51.4$ & $\textbf{70.5}$ & $77.7$ \\
        \rowcolor{gray!20} \textbf{DIG}~(\textit{Ours}) & \cellcolor{gray!20}16 & \cellcolor{gray!20}$\textbf{67.9}$ & \cellcolor{gray!20}$\textbf{78.0}$ & \cellcolor{gray!20}$\textbf{35.0}$ & \cellcolor{gray!20}$\textbf{66.8}$ & \cellcolor{gray!20}$\textbf{57.1}$ & \cellcolor{gray!20}$69.0$ & \cellcolor{gray!20}$\textbf{86.7}$ \\
        \midrule
        UNI & 32 & $61.8$ & $67.9$ & $18.5$ & $58.7$ & $57.4$ & $\textbf{76.0}$ & $86.7$ \\
        AKS~\cite{tang2025adaptivekeyframesamplinglong} & 32 & $66.8$ & $73.0$ & $40.3$ & $56.0$ & $\textbf{59.9}$ & $74.5$ & $\textbf{90.1}$ \\
        Q-Frame~\cite{zhang2025qframequeryawareframeselection} & 32 & $61.4$ & $67.9$ & $18.9$ & $63.8$ & $53.1$ & $71.5$ & $83.3$ \\
        \rowcolor{gray!20} \textbf{DIG}~(\textit{Ours}) & \cellcolor{gray!20}32 & \cellcolor{gray!20}$\textbf{72.4}$ & \cellcolor{gray!20}$\textbf{79.2}$ & \cellcolor{gray!20}$\textbf{48.1}$ & \cellcolor{gray!20}$\textbf{75.7}$ & \cellcolor{gray!20}$59.1$ & \cellcolor{gray!20}$74.0$ & \cellcolor{gray!20}$87.5$ \\
        \midrule
        UNI & 64 & $68.5$ & $72.1$ & $25.7$ & $61.4$ & $61.1$ & $\textbf{80.0}$ & $86.7$ \\
        AKS~\cite{tang2025adaptivekeyframesamplinglong} & 64 & $73.8$ & $76.6$ & $40.8$ & $58.3$ & $63.1$ & $75.0$ & $88.2$ \\
        Q-Frame~\cite{zhang2025qframequeryawareframeselection} & 64 & $68.5$ & $73.2$ & $21.8$ & $66.0$ & $59.7$ & $77.5$ & $85.9$ \\
        \rowcolor{gray!20} \textbf{DIG}~(\textit{Ours}) & \cellcolor{gray!20}64 & \cellcolor{gray!20}$\textbf{75.9}$ & \cellcolor{gray!20}$\textbf{81.1}$ & \cellcolor{gray!20}$\textbf{49.5}$ & \cellcolor{gray!20}$\textbf{78.4}$ & \cellcolor{gray!20}$\textbf{66.5}$ & \cellcolor{gray!20}$78.5$ & \cellcolor{gray!20}$\textbf{89.7}$ \\
        \midrule
        UNI & 128 & $73.5$ & $76.3$ & $30.6$ & $68.7$ & $64.2$ & $79.0$ & $89.4$ \\
        AKS~\cite{tang2025adaptivekeyframesamplinglong} & 128 & $78.3$ & $\textbf{80.3}$ & $42.2$ & $61.8$ & $\textbf{69.0}$ & $77.0$ & $87.8$ \\
        Q-Frame~\cite{zhang2025qframequeryawareframeselection} & 128 & $73.1$ & $76.1$ & $30.1$ & $69.1$ & $64.5$ & $\textbf{79.5}$ & $88.6$ \\
        \rowcolor{gray!20} \textbf{DIG}~(\textit{Ours}) & \cellcolor{gray!20}128 & \cellcolor{gray!20}$\textbf{79.8}$ & \cellcolor{gray!20}$80.0$ & \cellcolor{gray!20}$\textbf{52.4}$ & \cellcolor{gray!20}$\textbf{79.2}$ & \cellcolor{gray!20}$65.6$ & \cellcolor{gray!20}$78.5$ & \cellcolor{gray!20}$\textbf{89.7}$ \\
        \midrule
        UNI & 192 & $75.0$ & $78.0$ & $36.9$ & $69.9$ & $64.5$ & $78.0$ & $90.9$ \\
        AKS~\cite{tang2025adaptivekeyframesamplinglong} & 192 & $77.6$ & $\textbf{81.4}$ & $47.1$ & $63.3$ & $\textbf{68.2}$ & $77.0$ & $89.4$ \\
        \rowcolor{gray!20} \textbf{DIG}~(\textit{Ours}) & \cellcolor{gray!20}192 & \cellcolor{gray!20}$\textbf{82.6}$ & \cellcolor{gray!20}$\textbf{81.4}$ & \cellcolor{gray!20}$\textbf{53.9}$ & \cellcolor{gray!20}$\textbf{80.7}$ & \cellcolor{gray!20}$65.3$ & \cellcolor{gray!20}$\textbf{79.0}$ & \cellcolor{gray!20}$\textbf{91.6}$ \\
        \bottomrule
    \end{tabular}
    }
     \end{minipage}
     \hfill
     \begin{minipage}{0.49\linewidth} 
         \resizebox{\linewidth}{!}{
        \begin{tabular}{l|c|ccccccc}
        \toprule
        \multirow{2}{*}{Model} & \multirow{2}{*}{\#Frames} &  \multicolumn{7}{c}{MLVU~\cite{zhou2025mlvubenchmarkingmultitasklong}} \\ 
        \cmidrule(lr){3-9}
        & & PQA & NQA & AC & AO & ER & AR & TR\\
        \midrule
        UNI & 8 & $52.1$ & $62.5$ & $19.4$ & $44.0$ & $48.6$ & $66.5$ & $\textbf{82.9}$ \\
        Q-Frame~\cite{zhang2025qframequeryawareframeselection} & 8 & $50.6$ & $67.0$ & $19.9$ & $\textbf{48.6}$ & $49.7$ & $\textbf{68.0}$ & $73.8$ \\
        \rowcolor{gray!20} \textbf{DIG}~(\textit{Ours}) & \cellcolor{gray!20}8 & \cellcolor{gray!20}$\textbf{57.1}$ & \cellcolor{gray!20}$\textbf{73.2}$ & \cellcolor{gray!20}$\textbf{31.6}$ & \cellcolor{gray!20}$\textbf{48.6}$ & \cellcolor{gray!20}$\textbf{51.1}$ & \cellcolor{gray!20}$66.5$ & \cellcolor{gray!20}$\textbf{82.9}$ \\
        \midrule
        UNI & 16 & $56.0$ & $63.4$ & $19.9$ & $42.5$ & $54.8$ & $70.0$ & $84.4$ \\
        Q-Frame~\cite{zhang2025qframequeryawareframeselection} & 16 & $55.5$ & $67.6$ & $20.4$ & $50.6$ & $49.7$ & $\textbf{70.5}$ & $78.7$ \\
        \rowcolor{gray!20} \textbf{DIG}~(\textit{Ours}) & \cellcolor{gray!20}16 & \cellcolor{gray!20}$\textbf{66.4}$ & \cellcolor{gray!20}$\textbf{79.7}$ & \cellcolor{gray!20}$\textbf{36.9}$ & \cellcolor{gray!20}$\textbf{51.7}$ & \cellcolor{gray!20}$\textbf{55.4}$ & \cellcolor{gray!20}$68.5$ & \cellcolor{gray!20}$\textbf{85.2}$ \\
        \midrule
        UNI & 32 & $59.7$ & $69.0$ & $22.8$ & $51.4$ & $54.0$ & $\textbf{74.5}$ & $84.4$ \\
        AKS~\cite{tang2025adaptivekeyframesamplinglong} & 32 & $69.6$ & $76.3$ & $\textbf{42.2}$ & $50.2$ & $56.5$ & $72.0$ & $85.2$ \\
        Q-Frame~\cite{zhang2025qframequeryawareframeselection} & 32 & $60.1$ & $67.9$ & $23.9$ & $54.1$ & $54.3$ & $70.0$ & $83.7$ \\
        \rowcolor{gray!20} \textbf{DIG}~(\textit{Ours}) & \cellcolor{gray!20}32 & \cellcolor{gray!20}$\textbf{70.3}$ & \cellcolor{gray!20}$\textbf{80.6}$ & \cellcolor{gray!20}$\textbf{42.2}$ & \cellcolor{gray!20}$\textbf{54.4}$ & \cellcolor{gray!20}$\textbf{59.1}$ & \cellcolor{gray!20}$73.0$ & \cellcolor{gray!20}$\textbf{87.5}$ \\
        \midrule
        UNI & 64 & $64.0$ & $74.4$ & $26.2$ & $51.7$ & $59.9$ & $\textbf{76.0}$ & $87.1$ \\
        AKS~\cite{tang2025adaptivekeyframesamplinglong} & 64 & $69.9$ & $80.8$ & $41.3$ & $54.1$ & $60.5$ & $69.5$ & $84.8$ \\
        Q-Frame~\cite{zhang2025qframequeryawareframeselection} & 64 & $63.8$ & $73.5$ & $26.2$ & $56.0$ & $58.2$ & $72.0$ & $85.9$ \\
        \rowcolor{gray!20} \textbf{DIG}~(\textit{Ours}) & \cellcolor{gray!20}64 & \cellcolor{gray!20}$\textbf{75.3}$ & \cellcolor{gray!20}$\textbf{82.8}$ & \cellcolor{gray!20}$\textbf{46.6}$ & \cellcolor{gray!20}$\textbf{60.2}$ & \cellcolor{gray!20}$\textbf{62.2}$ & \cellcolor{gray!20}$75.5$ & \cellcolor{gray!20}$\textbf{87.8}$ \\
        \midrule
        UNI & 128 & $71.4$ & $79.2$ & $34.5$ & $58.7$ & $61.4$ & $73.0$ & $86.7$ \\
        AKS~\cite{tang2025adaptivekeyframesamplinglong} & 128 & $71.6$ & $\textbf{83.7}$ & $\textbf{48.5}$ & $57.1$ & $60.8$ & $72.0$ & $84.0$ \\
        Q-Frame~\cite{zhang2025qframequeryawareframeselection} & 128 & $71.4$ & $79.2$ & $34.5$ & $58.7$ & $61.4$ & $\textbf{73.0}$ & $86.7$ \\
        \rowcolor{gray!20} \textbf{DIG}~(\textit{Ours}) & \cellcolor{gray!20}128 & \cellcolor{gray!20}$\textbf{78.3}$ & \cellcolor{gray!20}$82.3$ & \cellcolor{gray!20}$45.6$ & \cellcolor{gray!20}$\textbf{62.5}$ & \cellcolor{gray!20}$\textbf{63.6}$ & \cellcolor{gray!20}$72.0$ & \cellcolor{gray!20}$\textbf{87.8}$ \\
        \midrule
        UNI & 192 & $72.0$ & $80.8$ & $40.3$ & $61.4$ & $63.6$ & $\textbf{73.0}$ & $\textbf{87.5}$ \\
        AKS~\cite{tang2025adaptivekeyframesamplinglong} & 192 & $74.2$ & $83.1$ & $46.1$ & $58.7$ & $63.6$ & $\textbf{73.0}$ & $85.6$ \\
        \rowcolor{gray!20} \textbf{DIG}~(\textit{Ours}) & \cellcolor{gray!20}192 & \cellcolor{gray!20}$\textbf{78.7}$ & \cellcolor{gray!20}$\textbf{84.5}$ & \cellcolor{gray!20}$\textbf{47.1}$ & \cellcolor{gray!20}$\textbf{63.3}$ & \cellcolor{gray!20}$\textbf{65.3}$ & \cellcolor{gray!20}$72.0$ & \cellcolor{gray!20}$\textbf{87.5}$ \\
        \midrule
        UNI & 256 & $73.5$ & $80.0$ & $41.3$ & $61.4$ & $61.1$ & $73.0$ & $\textbf{89.0}$ \\
        AKS~\cite{tang2025adaptivekeyframesamplinglong} & 256 & $75.3$ & $84.2$ & $47.3$ & $59.5$ & $\textbf{66.2}$ & $\textbf{75.5}$ & $87.8$ \\
        \rowcolor{gray!20} \textbf{DIG}~(\textit{Ours}) & \cellcolor{gray!20}256 & \cellcolor{gray!20}$\textbf{78.1}$ & \cellcolor{gray!20}$\textbf{84.5}$ & \cellcolor{gray!20}$\textbf{49.0}$ & \cellcolor{gray!20}$\textbf{62.2}$ & \cellcolor{gray!20}$65.1$ & \cellcolor{gray!20}$73.0$ & \cellcolor{gray!20}$\textbf{89.0}$ \\
        \bottomrule
    \end{tabular}
    }
     \end{minipage}
     
\end{table}

\begin{table}[t] 
  \caption{\textbf{Performance comparison between different frame selection methods on VideoMME.} Base LMMs are Qwen2.5-VL-7B~\cite{bai2025qwen25vltechnicalreport}(left) and Qwen2.5-VL-32B~\cite{bai2025qwen25vltechnicalreport}(right). \textbf{Bold} indicates best performance. The tasks are Object Reasoning~(ORA), Object Recognition~(ORC), Action Reasoning~(ARA), Information Synopsis~(INS), Counting Problem~(COP), Temporal Reasoning~(TER), Temporal Perception~(TEP), Spatial Perception~(SPP), Spatial Reasoning~(SPR), OCR, Attribute Perception~(ATP), Action Recognition~(ACR).} 
  \vspace{-3pt}
  \label{tab:exp_videomme_detail} 
  \centering 
  \begin{minipage}{0.49\linewidth} 
    \resizebox{\linewidth}{!}{
    \begin{tabular}{l|c|cccccccccccc}
         \toprule
         \multirow{2}{*}{Model} & \multirow{2}{*}{\#Frames} & \multicolumn{12}{c}{VideoMME~\cite{fu2024videommefirstevercomprehensiveevaluation}} \\
         \cmidrule(lr){3-14}
         & & ORA & ORC & ARA & INS & COP & TER & TEP & SPR & SPP & OCR & ATP & ACR \\
         \midrule
         UNI & 8 & $49.5$ & $54.5$ & $49.5$ & $67.8$ & $36.2$ & $\textbf{40.7}$ & $47.3$ & $58.9$ & $63.0$ & $48.9$ & $62.2$ & $53.4$ \\
         Q-Frame~\cite{zhang2025qframequeryawareframeselection} & 8 & $\textbf{50.5}$ & $50.1$ & $50.8$ & $64.8$ & $36.9$ & $36.0$ & $43.3$ & $62.1$ & $33.3$ & $44.0$ & $57.0$ & $52.2$ \\
         \rowcolor{gray!20} \textbf{DIG}~(\textit{Ours}) & 8 & $47.6$ & $\textbf{60.2}$ & $\textbf{52.3}$ & $\textbf{70.0}$ & $\textbf{39.6}$ & $\textbf{40.7}$ & $\textbf{56.4}$ & $\textbf{64.3}$ & $\textbf{66.7}$ & $\textbf{55.4}$ & $\textbf{65.3}$ & $\textbf{55.9}$ \\
         \midrule
         UNI & 16 & $53.5$ & $58.8$ & $51.2$ & $\textbf{74.6}$ & $37.7$ & $\textbf{43.5}$ & $\textbf{63.6}$ & $64.3$ & $72.2$ & $54.7$ & $67.1$ & $\textbf{56.9}$ \\
         Q-Frame~\cite{zhang2025qframequeryawareframeselection} & 16 & $52.4$ & $52.2$ & $52.2$ & $66.0$ & $36.8$ & $36.1$ & $56.9$ & $62.1$ & $45.9$ & $48.8$ & $59.0$ & $49.0$ \\
         \rowcolor{gray!20} \textbf{DIG}~(\textit{Ours}) & 16 & $\textbf{57.9}$ & $\textbf{61.0}$ & $\textbf{54.4}$ & $\textbf{74.6}$ & $\textbf{41.4}$ & $\textbf{43.5}$ & $56.4$ & $\textbf{66.1}$ & $\textbf{75.9}$ & $\textbf{59.0}$ & $\textbf{71.2}$ & $56.2$ \\
         \midrule
         UNI & 32 & $57.5$ & $63.6$ & $55.8$ & $76.5$ & $42.5$ & $43.5$ & $\textbf{67.3}$ & $\textbf{76.8}$ & $\textbf{72.2}$ & $62.6$ & $\textbf{74.8}$ & $59.7$ \\
         AKS~\cite{tang2025adaptivekeyframesamplinglong} & 32 & $56.8$ & $66.7$ & $51.9$ & $\textbf{80.2}$ & $\textbf{42.9}$ & $\textbf{49.2}$ & $60.0$ & $\textbf{76.8}$ & $\textbf{72.2}$ & $66.2$ & $\textbf{74.8}$ & $59.7$ \\
         Q-Frame~\cite{zhang2025qframequeryawareframeselection} & 32 & $55.3$ & $56.6$ & $54.2$ & $68.9$ & $38.3$ & $37.2$ & $59.6$ & $65.5$ & $50.1$ & $50.1$ & $64.0$ & $50.6$ \\
         \rowcolor{gray!20} \textbf{DIG}~(\textit{Ours}) & 32 & $\textbf{59.5}$ & $\textbf{67.2}$ & $\textbf{56.1}$ & $75.9$ & $40.3$ & $46.9$ & $52.7$ & $\textbf{76.8}$ & $\textbf{72.2}$ & $\textbf{69.8}$ & $73.4$ & $\textbf{61.0}$ \\
         \midrule
         UNI & 64 & $58.1$ & $66.7$ & $54.0$ & $76.8$ & $41.4$ & $43.5$ & $69.1$ & $\textbf{76.8}$ & $68.5$ & $67.6$ & $\textbf{74.8}$ & $\textbf{63.9}$ \\
         AKS~\cite{tang2025adaptivekeyframesamplinglong} & 64 & $58.4$ & $67.5$ & $56.1$ & $\textbf{79.3}$ & $44.4$ & $\textbf{52.0}$ & $\textbf{72.7}$ & $\textbf{76.8}$ & $72.2$ & $72.7$ & $\textbf{74.8}$ & $61.0$ \\
         Q-Frame~\cite{zhang2025qframequeryawareframeselection} & 64 & $56.0$ & $62.4$ & $56.0$ & $72.6$ & $\textbf{46.5}$ & $45.2$ & $56.8$ & $69.1$ & $54.2$ & $51.3$ & $72.0$ & $51.2$ \\
         \rowcolor{gray!20} \textbf{DIG}~(\textit{Ours}) & 64 & $\textbf{60.6}$ & $\textbf{68.4}$ & $\textbf{57.5}$ & $78.0$ & $46.3$ & $49.2$ & $58.2$ & $73.2$ & $\textbf{75.9}$ & $\textbf{73.4}$ & $73.4$ & $63.3$ \\
         \midrule
         UNI & 128 & $61.2$ & $71.2$ & $58.9$ & $79.9$ & $\textbf{45.1}$ & $\textbf{57.1}$ & $70.9$ & $76.8$ & $\textbf{68.5}$ & $71.2$ & $76.6$ & $66.1$ \\
         AKS~\cite{tang2025adaptivekeyframesamplinglong} & 128 & $61.0$ & $68.4$ & $\textbf{59.3}$ & $80.2$ & $44.8$ & $\textbf{57.1}$ & $\textbf{76.4}$ & $75.0$ & $\textbf{68.5}$ & $\textbf{77.0}$ & $77.9$ & $61.3$ \\
         Q-Frame~\cite{zhang2025qframequeryawareframeselection} & 128 & $58.7$ & $64.5$ & $55.6$ & $78.0$ & $38.9$ & $55.6$ & $64.9$ & $72.4$ & $54.3$ & $57.4$ & $67.0$ & $59.3$ \\
         \rowcolor{gray!20} \textbf{DIG}~(\textit{Ours}) & 128 & $\textbf{61.7}$ & $\textbf{72.3}$ & $57.2$ & $\textbf{80.5}$ & $\textbf{45.1}$ & $52.5$ & $58.2$ & $\textbf{78.6}$ & $\textbf{68.5}$ & $71.2$ & $\textbf{78.8}$ & $\textbf{66.8}$ \\
         \midrule
         UNI & 192 & $62.3$ & $72.0$ & $\textbf{60.7}$ & $79.9$ & $\textbf{48.5}$ & $54.2$ & $72.7$ & $76.8$ & $70.4$ & $71.9$ & $77.5$ & $\textbf{68.1}$ \\
         AKS~\cite{tang2025adaptivekeyframesamplinglong} & 192 & $60.8$ & $69.2$ & $57.5$ & $79.3$ & $45.5$ & $\textbf{57.6}$ & $\textbf{81.8}$ & $76.8$ & $\textbf{72.2}$ & $\textbf{76.3}$ & $77.9$ & $63.9$ \\
         \rowcolor{gray!20} \textbf{DIG}~(\textit{Ours}) & 192 & $\textbf{64.5}$ & $\textbf{73.7}$ & $60.0$ & $\textbf{80.2}$ & $46.3$ & $54.2$ & $65.5$ & $\textbf{78.6}$ & $68.5$ & $74.1$ & $\textbf{79.3}$ & $65.8$ \\
         \midrule
         UNI & 256 & $62.1$ & $71.5$ & $59.6$ & $\textbf{82.4}$ & $\textbf{46.6}$ & $\textbf{57.6}$ & $76.4$ & $75.0$ & $66.7$ & $71.9$ & $77.9$ & $\textbf{68.1}$ \\
         AKS~\cite{tang2025adaptivekeyframesamplinglong} & 256 & $61.0$ & $70.9$ & $57.9$ & $79.3$ & $44.8$ & $57.1$ & $\textbf{83.6}$ & $75.0$ & $\textbf{70.4}$ & $\textbf{74.1}$ & $77.9$ & $64.5$ \\
         \rowcolor{gray!20} \textbf{DIG}~(\textit{Ours}) & 256 & $\textbf{63.0}$ & $\textbf{72.0}$ & $\textbf{62.1}$ & $\textbf{82.4}$ & $46.3$ & $54.8$ & $67.3$ & $\textbf{78.6}$ & $66.7$ & $73.4$ & $\textbf{80.2}$ & $67.4$ \\
         \bottomrule
    \end{tabular}
    }
    \end{minipage}
    \hfill
    \begin{minipage}{0.49\linewidth} 
    \resizebox{\linewidth}{!}{
      \begin{tabular}{l|c|cccccccccccc}
         \toprule
         \multirow{2}{*}{Model} & \multirow{2}{*}{\#Frames} & \multicolumn{12}{c}{VideoMME~\cite{fu2024videommefirstevercomprehensiveevaluation}} \\ 
         \cmidrule(lr){3-14}
         & & ORA & ORC & ARA & INS & COP & TER & TEP & SPR & SPP & OCR & ATP & ACR \\
         \midrule
        UNI & 8 & $\textbf{57.2}$ & $54.8$ & $\textbf{53.4}$ & $\textbf{69.7}$ & $30.1$ & $37.8$ & $\textbf{46.0}$ & $65.5$ & $54.2$ & $47.6$ & $\textbf{62.0}$ & $45.6$ \\
         Q-Frame~\cite{zhang2025qframequeryawareframeselection} & 8 & $50.5$ & $50.1$ & $50.8$ & $64.8$ & $\textbf{36.9}$ & $36.0$ & $43.3$ & $62.1$ & $33.3$ & $44.0$ & $57.0$ & $\textbf{52.2}$ \\
         \rowcolor{gray!20} \textbf{DIG}~(\textit{Ours}) & 8 & $55.6$ & $\textbf{58.1}$ & $\textbf{53.4}$ & $\textbf{69.7}$ & $31.5$ & $\textbf{40.9}$ & $\textbf{46.0}$ & $\textbf{72.4}$ & $\textbf{62.5}$ & $\textbf{54.9}$ & $\textbf{62.0}$ & $48.4$ \\
         \midrule
         UNI & 16 & $55.4$ & $56.5$ & $52.5$ & $70.5$ & $32.2$ & $\textbf{47.6}$ & $51.4$ & $\textbf{75.9}$ & $\textbf{62.5}$ & $42.7$ & $\textbf{65.0}$ & $\textbf{50.6}$ \\
         Q-Frame~\cite{zhang2025qframequeryawareframeselection} & 16 & $52.4$ & $52.2$ & $52.2$ & $66.0$ & $36.8$ & $36.1$ & $\textbf{56.9}$ & $62.1$ & $45.9$ & $\textbf{48.8}$ & $59.0$ & $49.0$ \\
         \rowcolor{gray!20} \textbf{DIG}~(\textit{Ours}) & 16 & $\textbf{58.3}$ & $\textbf{59.7}$ & $\textbf{53.4}$ & $\textbf{71.0}$ & $\textbf{37.8}$ & $47.0$ & $56.8$ & $\textbf{75.9}$ & $\textbf{62.5}$ & $45.1$ & $\textbf{65.0}$ & $47.3$ \\
         \midrule
         UNI & 32 & $55.4$ & $55.9$ & $54.2$ & $76.8$ & $35.0$ & $40.9$ & $\textbf{62.2}$ & $75.9$ & $\textbf{62.5}$ & $47.6$ & $66.0$ & $\textbf{51.7}$ \\
         AKS~\cite{tang2025adaptivekeyframesamplinglong} & 32 & $58.6$ & $57.0$ & $\textbf{57.6}$ & $\textbf{78.0}$ & $34.3$ & $47.6$ & $56.8$ & $\textbf{79.3}$ & $45.8$ & $\textbf{61.0}$ & $68.0$ & $51.1$ \\
         Q-Frame~\cite{zhang2025qframequeryawareframeselection} & 32 & $55.3$ & $56.6$ & $54.2$ & $68.9$ & $\textbf{38.3}$ & $37.2$ & $59.6$ & $65.5$ & $50.1$ & $50.1$ & $64.0$ & $50.6$ \\
         \rowcolor{gray!20} \textbf{DIG}~(\textit{Ours}) & 32 & $\textbf{58.8}$ & $\textbf{64.0}$ & $55.5$ & $\textbf{78.0}$ & $35.7$ & $\textbf{51.8}$ & $48.7$ & $75.9$ & $58.3$ & $56.1$ & $\textbf{69.0}$ & $50.0$ \\
         \midrule
         UNI & 64 & $61.8$ & $62.9$ & $58.0$ & $76.8$ & $34.3$ & $44.5$ & $\textbf{64.9}$ & $\textbf{79.3}$ & $\textbf{62.5}$ & $58.5$ & $69.0$ & $\textbf{59.3}$ \\
         AKS~\cite{tang2025adaptivekeyframesamplinglong} & 64 & $61.0$ & $63.4$ & $\textbf{59.7}$ & $\textbf{80.1}$ & $37.8$ & $\textbf{54.3}$ & $62.2$ & $\textbf{79.3}$ & $54.2$ & $59.8$ & $\textbf{73.0}$ & $56.6$ \\
         Q-Frame~\cite{zhang2025qframequeryawareframeselection} & 64 & $56.0$ & $62.4$ & $56.0$ & $72.6$ & $\textbf{46.5}$ & $45.2$ & $56.8$ & $69.1$ & $54.2$ & $51.3$ & $72.0$ & $51.2$ \\
         \rowcolor{gray!20} \textbf{DIG}~(\textit{Ours}) & 64 & $\textbf{62.6}$ & $\textbf{65.6}$ & $57.6$ & $76.4$ & $32.9$ & $51.8$ & $59.5$ & $75.9$ & $58.3$ & $\textbf{67.1}$ & $\textbf{73.0}$ & $58.2$ \\
         \midrule
         UNI & 128 & $63.4$ & $69.9$ & $\textbf{63.5}$ & $81.3$ & $42.0$ & $54.3$ & $59.5$ & $\textbf{86.2}$ & $\textbf{58.3}$ & $68.3$ & $73.0$ & $57.1$ \\
         AKS~\cite{tang2025adaptivekeyframesamplinglong} & 128 & $\textbf{66.0}$ & $68.3$ & $58.8$ & $81.3$ & $42.7$ & $\textbf{57.9}$ & $\textbf{67.6}$ & $82.8$ & $50.0$ & $\textbf{69.5}$ & $\textbf{75.0}$ & $59.9$ \\
         Q-Frame~\cite{zhang2025qframequeryawareframeselection} & 128 & $58.7$ & $64.5$ & $55.6$ & $78.0$ & $38.9$ & $55.6$ & $64.9$ & $72.4$ & $54.3$ & $57.4$ & $67.0$ & $59.3$ \\
         \rowcolor{gray!20} \textbf{DIG}~(\textit{Ours}) & 128 & $65.2$ & $\textbf{73.1}$ & $61.8$ & $\textbf{81.7}$ & $\textbf{45.5}$ & $52.4$ & $64.9$ & $82.8$ & $54.2$ & $\textbf{69.5}$ & $\textbf{75.0}$ & $\textbf{60.4}$ \\
         \midrule
         UNI & 192 & $64.7$ & $69.9$ & $\textbf{63.0}$ & $81.3$ & $\textbf{44.8}$ & $55.5$ & $73.0$ & $\textbf{86.2}$ & $\textbf{62.5}$ & $68.3$ & $75.0$ & $\textbf{62.1}$ \\
         AKS~\cite{tang2025adaptivekeyframesamplinglong} & 192 & $65.5$ & $69.4$ & $59.7$ & $81.3$ & $42.7$ & $\textbf{58.5}$ & $\textbf{75.7}$ & $79.3$ & $54.2$ & $\textbf{73.2}$ & $78.0$ & $58.8$ \\
         \rowcolor{gray!20} \textbf{DIG}~(\textit{Ours}) & 192 & $\textbf{66.8}$ & $\textbf{74.2}$ & $62.6$ & $\textbf{81.7}$ & $42.0$ & $57.3$ & $64.9$ & $79.3$ & $58.3$ & $\textbf{73.2}$ & $\textbf{80.0}$ & $61.0$ \\
         \bottomrule
    \end{tabular}
    }
    \end{minipage}
    \vspace{-3pt}
\end{table}

\begin{table}[t]
    \caption{\textbf{Performance Comparison between Different Frame Selection Methods on LongVideoBench.} Base LMMs are Qwen2.5-VL-7B~\cite{bai2025qwen25vltechnicalreport}(top) and Qwen2.5-VL-32B~\cite{bai2025qwen25vltechnicalreport}(bottom). \textbf{Bold} indicates best performance.} 
    \label{tab:exp_lv_detail}
    \centering
    \small
    \resizebox{0.85\linewidth}{!}{
    \begin{tabular}{l|c|ccccccccc|cccccccccc}
        \toprule
        \multirow{3}{*}{Model} & \multirow{3}{*}{\#Frames} & \multicolumn{19}{c}{LongVideoBench~\cite{wu2024longvideobenchbenchmarklongcontextinterleaved}} \\
        \cmidrule(lr){3-21}
        & & \multicolumn{9}{c|}{L1-Perception} & \multicolumn{10}{c}{L2-Relation} \\
        \cmidrule(lr){3-21}
        & & S2E & S2A & O2E & T2O & S2O & T2E & E2O & T2A & \cellcolor{gray!20}Avg & TOS & E3E & SAA & O3O & T3O & T3E & TAA & SSS & SOS & \cellcolor{gray!20}Avg \\
        \midrule
        UNI & 8 & $57.0$ & $51.1$ & $\textbf{62.8}$ & $56.6$ & $45.8$ & $58.5$ & $56.9$ & $49.4$ & \cellcolor{gray!20}$54.4$ & $\textbf{38.4}$ & $\textbf{62.8}$ & $47.2$ & $45.5$ & $47.3$ & $47.9$ & $\textbf{46.3}$ & $\textbf{34.0}$ & $64.2$ & \cellcolor{gray!20}$48.3$ \\
        Q-Frame~\cite{zhang2025qframequeryawareframeselection} & 8 & $67.7$ & $\textbf{73.9}$ & $60.9$ & $\textbf{57.9}$ & $55.6$ & $\textbf{64.6}$ & $\textbf{63.1}$ & $55.7$ & \cellcolor{gray!20}$\textbf{62.7}$ & $31.5$ & $62.8$ & $\textbf{52.8}$ & $47.0$ & $39.2$ & $\textbf{48.0}$ & $45.1$ & $28.9$ & $65.4$ & \cellcolor{gray!20}$46.8$ \\
        \rowcolor{gray!20} \rowcolor{gray!20} \textbf{DIG}~(\textit{Ours}) & 8 & $\textbf{69.9}$ & $68.2$ & $62.1$ & $50.0$ & $\textbf{55.6}$ & $61.5$ & $61.5$ & $\textbf{62.0}$ & $61.8$ & $37.0$ & $61.7$ & $50.0$ & $\textbf{48.5}$ & $\textbf{54.1}$ & $43.8$ & $45.1$ & $29.9$ & $\textbf{67.9}$ & $\textbf{48.6}$ \\
        \midrule
        UNI & 16 & $65.6$ & $64.8$ & $62.8$ & $53.9$ & $48.6$ & $61.5$ & $61.5$ & $51.9$ & \cellcolor{gray!20}$59.0$ & $\textbf{37.0}$ & $\textbf{62.8}$ & $50.0$ & $47.0$ & $\textbf{56.8}$ & $45.2$ & $\textbf{51.9}$ & $36.1$ & $63.0$ & \cellcolor{gray!20}$49.3$ \\
        Q-Frame~\cite{zhang2025qframequeryawareframeselection} & 16 & $66.7$ & $70.5$ & $\textbf{65.5}$ & $63.2$ & $\textbf{61.1}$ & $64.6$ & $\textbf{69.2}$ & $63.3$ & \cellcolor{gray!20}$\textbf{65.6}$ & $34.3$ & $59.6$ & $\textbf{56.9}$ & $\textbf{54.6}$ & $43.2$ & $49.3$ & $50.0$ & $\textbf{38.1}$ & $65.4$ & \cellcolor{gray!20}$50.1$ \\
        \rowcolor{gray!20} \rowcolor{gray!20} \textbf{DIG}~(\textit{Ours}) & 16 & $\textbf{72.0}$ & $\textbf{71.6}$ & $59.8$ & $\textbf{65.8}$ & $54.2$ & $\textbf{69.2}$ & $\textbf{69.2}$ & $\textbf{63.3}$ & $65.4$ & $\textbf{37.0}$ & $57.4$ & $52.8$ & $43.9$ & $\textbf{56.8}$ & $\textbf{52.1}$ & $43.9$ & $36.1$ & $\textbf{74.1}$ & $\textbf{50.4}$ \\
        \midrule
        UNI & 32 & $67.7$ & $58.0$ & $61.7$ & $56.6$ & $62.5$ & $\textbf{67.7}$ & $67.7$ & $51.9$ & \cellcolor{gray!20}$61.9$ & $37.0$ & $61.7$ & $55.6$ & $\textbf{56.1}$ & $\textbf{55.4}$ & $49.3$ & $\textbf{52.4}$ & $\textbf{40.2}$ & $66.7$ & \cellcolor{gray!20}$52.7$ \\
        AKS~\cite{tang2025adaptivekeyframesamplinglong} & 32 & $65.6$ & $77.3$ & $\textbf{67.8}$ & $63.2$ & $\textbf{63.9}$ & $63.1$ & $63.1$ & $\textbf{64.6}$ & \cellcolor{gray!20}$66.1$ & $37.0$ & $\textbf{67.0}$ & $58.3$ & $\textbf{56.1}$ & $45.9$ & $\textbf{53.4}$ & $48.8$ & $38.1$ & $\textbf{74.1}$ & \cellcolor{gray!20}$\textbf{53.2}$ \\
        Q-Frame~\cite{zhang2025qframequeryawareframeselection} & 32 & $64.5$ & $69.3$ & $60.9$ & $59.2$ & $61.1$ & $61.5$ & $\textbf{69.2}$ & $60.8$ & \cellcolor{gray!20}$63.3$ & $32.9$ & $59.6$ & $\textbf{62.5}$ & $50.0$ & $50.0$ & $45.2$ & $46.3$ & $39.2$ & $66.7$ & \cellcolor{gray!20}$50.3$ \\
        \rowcolor{gray!20} \rowcolor{gray!20} \textbf{DIG}~(\textit{Ours}) & 32 & $\textbf{72.0}$ & $\textbf{78.4}$ & $63.2$ & $\textbf{68.4}$ & $62.5$ & $\textbf{67.7}$ & $67.7$ & $62.0$ & $\textbf{68.0}$ & $\textbf{41.1}$ & $62.8$ & $52.8$ & $51.5$ & $\textbf{55.4}$ & $50.7$ & $48.8$ & $36.1$ & $70.4$ & $52.1$ \\
        \midrule
        UNI & 64 & $\textbf{73.1}$ & $67.0$ & $62.8$ & $59.2$ & $56.9$ & $63.1$ & $66.2$ & $62.0$ & \cellcolor{gray!20}$64.6$ & $34.2$ & $62.8$ & $58.3$ & $59.1$ & $\textbf{60.8}$ & $47.9$ & $51.2$ & $\textbf{43.3}$ & $67.9$ & \cellcolor{gray!20}$53.9$ \\
        AKS~\cite{tang2025adaptivekeyframesamplinglong} & 64 & $69.9$ & $77.3$ & $\textbf{71.3}$ & $65.8$ & $59.7$ & $60.0$ & $64.6$ & $64.6$ & \cellcolor{gray!20}$67.2$ & $\textbf{41.1}$ & $\textbf{70.2}$ & $\textbf{61.1}$ & $56.1$ & $47.3$ & $49.3$ & $47.6$ & $40.2$ & $\textbf{76.5}$ & \cellcolor{gray!20}$54.5$ \\
        Q-Frame~\cite{zhang2025qframequeryawareframeselection} & 64 & $63.4$ & $70.5$ & $63.2$ & $57.9$ & $\textbf{61.1}$ & $63.1$ & $64.6$ & $65.8$ & \cellcolor{gray!20}$63.8$ & $34.3$ & $67.0$ & $58.3$ & $53.0$ & $52.7$ & $46.6$ & $\textbf{51.2}$ & $37.1$ & $66.7$ & \cellcolor{gray!20}$52.0$ \\
        \rowcolor{gray!20} \rowcolor{gray!20} \textbf{DIG}~(\textit{Ours}) & 64 & $69.9$ & $\textbf{78.4}$ & $66.7$ & $\textbf{69.7}$ & $55.6$ & $\textbf{67.7}$ & $\textbf{72.3}$ & $\textbf{68.4}$ & $\textbf{68.8}$ & $38.4$ & $66.0$ & $58.3$ & $\textbf{62.1}$ & $59.5$ & $\textbf{53.4}$ & $46.3$ & $42.3$ & $69.1$ & $\textbf{54.9}$ \\
        \midrule
        UNI & 128 & $71.0$ & $67.0$ & $67.8$ & $64.5$ & $\textbf{61.1}$ & $67.7$ & $72.3$ & $\textbf{73.4}$ & \cellcolor{gray!20}$68.2$ & $\textbf{38.4}$ & $\textbf{68.1}$ & $56.9$ & $59.1$ & $\textbf{56.8}$ & $50.7$ & $\textbf{54.9}$ & $46.4$ & $74.1$ & \cellcolor{gray!20}$\textbf{56.3}$ \\
        AKS~\cite{tang2025adaptivekeyframesamplinglong} & 128 & $69.9$ & $72.7$ & $66.7$ & $64.5$ & $\textbf{61.1}$ & $60.0$ & $66.2$ & $64.6$ & \cellcolor{gray!20}$66.1$ & $37.0$ & $\textbf{68.1}$ & $\textbf{63.9}$ & $56.1$ & $50.0$ & $50.7$ & $48.8$ & $\textbf{47.4}$ & $\textbf{76.5}$ & \cellcolor{gray!20}$55.6$ \\
        Q-Frame~\cite{zhang2025qframequeryawareframeselection} & 128 & $66.7$ & $67.1$ & $\textbf{69.0}$ & $60.5$ & $59.7$ & $64.6$ & $67.7$ & $64.6$ & \cellcolor{gray!20}$65.1$ & $\textbf{38.4}$ & $64.9$ & $58.3$ & $54.6$ & $\textbf{56.8}$ & $49.3$ & $51.2$ & $45.4$ & $75.3$ & \cellcolor{gray!20}$55.1$ \\
        \rowcolor{gray!20} \rowcolor{gray!20} \textbf{DIG}~(\textit{Ours}) & 128 & $\textbf{72.0}$ & $\textbf{79.5}$ & $66.7$ & $\textbf{71.1}$ & $\textbf{61.1}$ & $\textbf{69.2}$ & $\textbf{75.4}$ & $70.9$ & $\textbf{70.9}$ & $\textbf{38.4}$ & $\textbf{68.1}$ & $61.1$ & $\textbf{65.2}$ & $\textbf{56.8}$ & $\textbf{57.5}$ & $47.6$ & $45.4$ & $67.9$ & $\textbf{56.3}$ \\
        \midrule
        UNI & 192 & $\textbf{74.2}$ & $72.7$ & $66.0$ & $\textbf{68.4}$ & $59.7$ & $67.7$ & $70.8$ & $63.3$ & \cellcolor{gray!20}$68.2$ & $\textbf{35.6}$ & $66.0$ & $59.7$ & $59.1$ & $\textbf{58.1}$ & $58.9$ & $\textbf{56.1}$ & $46.4$ & $67.9$ & \cellcolor{gray!20}$56.5$ \\
        AKS~\cite{tang2025adaptivekeyframesamplinglong} & 192 & $69.9$ & $76.1$ & $\textbf{67.8}$ & $63.2$ & $\textbf{61.1}$ & $63.1$ & $67.7$ & $65.8$ & \cellcolor{gray!20}$67.2$ & $\textbf{35.6}$ & $\textbf{68.1}$ & $\textbf{62.5}$ & $56.1$ & $56.8$ & $52.1$ & $48.8$ & $46.4$ & $\textbf{72.8}$ & \cellcolor{gray!20}$55.6$ \\
        \rowcolor{gray!20} \rowcolor{gray!20} \textbf{DIG}~(\textit{Ours}) & 192 & $\textbf{74.2}$ & $\textbf{84.1}$ & $66.7$ & $67.1$ & $59.7$ & $\textbf{69.2}$ & $\textbf{73.8}$ & $\textbf{78.5}$ & $\textbf{72.0}$ & $\textbf{35.6}$ & $\textbf{68.1}$ & $61.1$ & $\textbf{66.7}$ & $56.8$ & $\textbf{61.6}$ & $48.8$ & $\textbf{49.5}$ & $70.4$ & $\textbf{57.6}$ \\
        \midrule
        UNI & 256 & $69.9$ & $73.9$ & $69.1$ & $63.2$ & $59.7$ & $63.1$ & $72.3$ & $68.4$ & \cellcolor{gray!20}$66.7$ & $\textbf{37.0}$ & $69.1$ & $\textbf{62.5}$ & $60.6$ & $55.4$ & $\textbf{56.2}$ & $\textbf{54.9}$ & $46.4$ & $69.1$ & \cellcolor{gray!20}$\textbf{56.9}$ \\
        AKS~\cite{tang2025adaptivekeyframesamplinglong} & 256 & $68.8$ & $72.7$ & $70.1$ & $65.8$ & $61.1$ & $66.2$ & $63.1$ & $65.8$ & \cellcolor{gray!20}$67.0$ & $35.6$ & $69.1$ & $\textbf{62.5}$ & $56.1$ & $\textbf{56.8}$ & $52.1$ & $48.8$ & $43.3$ & $\textbf{71.6}$ & \cellcolor{gray!20}$55.2$ \\
        \rowcolor{gray!20} \rowcolor{gray!20} \textbf{DIG}~(\textit{Ours}) & 256 & $\textbf{76.3}$ & $\textbf{80.7}$ & $\textbf{71.3}$ & $\textbf{72.4}$ & $\textbf{65.3}$ & $\textbf{69.2}$ & $\textbf{75.4}$ & $\textbf{74.7}$ & $\textbf{73.4}$ & $\textbf{37.0}$ & $\textbf{71.3}$ & $59.7$ & $\textbf{68.2}$ & $\textbf{56.8}$ & $\textbf{56.2}$ & $45.1$ & $\textbf{49.5}$ & $67.9$ & $\textbf{56.9}$ \\
        \bottomrule
    \end{tabular}
    }\\
    \vspace{12pt}
    \resizebox{0.85\linewidth}{!}{
    \begin{tabular}{l|c|ccccccccc|cccccccccc}
        \toprule
        \multirow{3}{*}{Model} & \multirow{3}{*}{\#Frames} & \multicolumn{19}{c}{LongVideoBench~\cite{wu2024longvideobenchbenchmarklongcontextinterleaved}} \\
        \cmidrule(lr){3-21}
        & & \multicolumn{9}{c|}{L1-Perception} & \multicolumn{10}{c}{L2-Relation} \\ 
        \cmidrule(lr){3-21}
        & & S2E & S2A & O2E & T2O & S2O & T2E & E2O & T2A & \cellcolor{gray!20}Avg & TOS & E3E & SAA & O3O & T3O & T3E & TAA & SSS & SOS & \cellcolor{gray!20}Avg \\
        \midrule
        UNI & 8 & $62.4$ & $61.4$ & $\textbf{65.5}$ & $54.0$ & $51.4$ & $58.5$ & $58.5$ & $51.9$ & \cellcolor{gray!20}$58.2$ & $30.1$ & $\textbf{63.8}$ & $55.6$ & $42.4$ & $47.3$ & $\textbf{49.3}$ & $45.1$ & $\textbf{46.4}$ & $58.0$ & \cellcolor{gray!20}$49.2$ \\
        Q-Frame~\cite{zhang2025qframequeryawareframeselection} & 8 & $62.4$ & $\textbf{81.8}$ & $62.1$ & $55.3$ & $\textbf{55.6}$ & $\textbf{61.5}$ & $\textbf{66.2}$ & $53.2$ & \cellcolor{gray!20}$\textbf{62.6}$ & $30.1$ & $57.5$ & $56.9$ & $42.4$ & $43.2$ & $46.6$ & $41.5$ & $36.1$ & $59.3$ & \cellcolor{gray!20}$46.1$ \\
        \rowcolor{gray!20} \rowcolor{gray!20} \textbf{DIG}~(\textit{Ours}) & 8 & $\textbf{63.4}$ & $75.0$ & $60.9$ & $\textbf{59.2}$ & $48.6$ & $60.0$ & $66.2$ & $\textbf{60.8}$ & $61.8$ & $\textbf{37.0}$ & $60.6$ & $\textbf{61.1}$ & $\textbf{50.0}$ & $\textbf{52.7}$ & $\textbf{49.3}$ & $\textbf{47.6}$ & $44.3$ & $\textbf{65.4}$ & $\textbf{52.0}$ \\
        \midrule
        UNI & 16 & $57.0$ & $71.6$ & $57.5$ & $52.6$ & $48.6$ & $63.1$ & $63.1$ & $45.6$ & \cellcolor{gray!20}$57.4$ & $\textbf{37.0}$ & $58.5$ & $58.3$ & $53.0$ & $54.1$ & $50.7$ & $\textbf{53.7}$ & $\textbf{42.3}$ & $66.7$ & \cellcolor{gray!20}$\textbf{52.7}$ \\
        Q-Frame~\cite{zhang2025qframequeryawareframeselection} & 16 & $\textbf{69.9}$ & $77.3$ & $\textbf{64.4}$ & $\textbf{64.5}$ & $58.3$ & $64.6$ & $63.1$ & $62.0$ & \cellcolor{gray!20}$\textbf{65.9}$ & $31.5$ & $\textbf{62.8}$ & $58.3$ & $51.5$ & $46.0$ & $43.8$ & $46.3$ & $40.2$ & $54.3$ & \cellcolor{gray!20}$49.5$ \\
        \rowcolor{gray!20} \rowcolor{gray!20} \textbf{DIG}~(\textit{Ours}) & 16 & $66.7$ & $\textbf{80.7}$ & $63.2$ & $55.3$ & $\textbf{62.5}$ & $\textbf{66.2}$ & $\textbf{64.6}$ & $\textbf{68.4}$ & $\textbf{65.9}$ & $34.3$ & $56.4$ & $\textbf{63.9}$ & $\textbf{54.5}$ & $\textbf{56.8}$ & $\textbf{54.8}$ & $46.3$ & $39.2$ & $\textbf{67.9}$ & $52.7$ \\
        \midrule
        UNI & 32 & $68.8$ & $68.2$ & $64.4$ & $57.9$ & $54.2$ & $63.1$ & $61.5$ & $53.2$ & \cellcolor{gray!20}$61.8$ & $32.9$ & $66.0$ & $56.9$ & $53.0$ & $52.7$ & $\textbf{58.9}$ & $50.0$ & $47.4$ & $70.4$ & \cellcolor{gray!20}$\textbf{54.5}$ \\
        AKS~\cite{tang2025adaptivekeyframesamplinglong} & 32 & $65.6$ & $\textbf{77.3}$ & $\textbf{69.0}$ & $\textbf{61.8}$ & $\textbf{65.3}$ & $61.5$ & $66.2$ & $\textbf{64.6}$ & \cellcolor{gray!20}$\textbf{66.7}$ & $34.3$ & $69.2$ & $59.7$ & $50.0$ & $41.9$ & $48.0$ & $\textbf{51.2}$ & $44.3$ & $\textbf{72.8}$ & \cellcolor{gray!20}$52.8$ \\
        Q-Frame~\cite{zhang2025qframequeryawareframeselection} & 32 & $69.9$ & $\textbf{77.3}$ & $63.2$ & $59.2$ & $\textbf{65.3}$ & $61.5$ & $60.0$ & $54.4$ & \cellcolor{gray!20}$61.4$ & $\textbf{37.0}$ & $59.6$ & $54.2$ & $48.5$ & $48.6$ & $48.0$ & $\textbf{51.2}$ & $50.5$ & $60.5$ & \cellcolor{gray!20}$51.3$ \\
        \rowcolor{gray!20} \rowcolor{gray!20} \textbf{DIG}~(\textit{Ours}) & 32 & $\textbf{71.0}$ & $75.0$ & $65.5$ & $57.9$ & $\textbf{65.3}$ & $\textbf{72.3}$ & $\textbf{67.7}$ & $60.8$ & $66.9$ & $35.6$ & $\textbf{70.2}$ & $\textbf{66.7}$ & $\textbf{62.1}$ & $\textbf{56.8}$ & $56.2$ & $43.9$ & $\textbf{51.5}$ & $71.6$ & $\textbf{57.2}$ \\
        \midrule
        UNI & 64 & $\textbf{69.9}$ & $65.9$ & $64.4$ & $61.8$ & $58.3$ & $60.0$ & $70.8$ & $57.0$ & \cellcolor{gray!20}$63.7$ & $31.5$ & $67.0$ & $58.3$ & $51.5$ & $55.4$ & $\textbf{54.8}$ & $51.2$ & $51.5$ & $69.1$ & \cellcolor{gray!20}$54.9$ \\
        AKS~\cite{tang2025adaptivekeyframesamplinglong} & 64 & $67.7$ & $76.1$ & $\textbf{66.7}$ & $60.5$ & $\textbf{69.4}$ & $61.5$ & $\textbf{72.3}$ & $\textbf{67.1}$ & \cellcolor{gray!20}$67.8$ & $\textbf{37.0}$ & $\textbf{73.4}$ & $56.9$ & $53.0$ & $47.3$ & $52.1$ & $48.8$ & $52.6$ & $\textbf{75.3}$ & \cellcolor{gray!20}$\textbf{55.8}$ \\
        Q-Frame~\cite{zhang2025qframequeryawareframeselection} & 64 & $63.4$ & $77.3$ & $\textbf{66.7}$ & $56.6$ & $66.7$ & $64.6$ & $67.7$ & $65.8$ & \cellcolor{gray!20}$64.0$ & $35.6$ & $63.8$ & $61.1$ & $47.0$ & $46.0$ & $52.1$ & $\textbf{52.4}$ & $53.6$ & $67.9$ & \cellcolor{gray!20}$53.8$ \\
        \rowcolor{gray!20} \rowcolor{gray!20} \textbf{DIG}~(\textit{Ours}) & 64 & $\textbf{69.9}$ & $\textbf{79.5}$ & $65.5$ & $\textbf{64.5}$ & $65.3$ & $\textbf{66.2}$ & $\textbf{72.3}$ & $\textbf{67.1}$ & $\textbf{68.8}$ & $34.3$ & $68.1$ & $\textbf{73.6}$ & $\textbf{62.1}$ & $\textbf{59.5}$ & $\textbf{54.8}$ & $46.3$ & $\textbf{57.7}$ & $72.8$ & $\textbf{55.8}$ \\
        \midrule
        UNI & 128 & $71.0$ & $70.5$ & $62.1$ & $61.8$ & $68.1$ & $63.1$ & $67.7$ & $60.8$ & \cellcolor{gray!20}$65.7$ & $35.6$ & $70.2$ & $62.5$ & $51.5$ & $55.4$ & $57.5$ & $\textbf{53.7}$ & $\textbf{60.8}$ & $71.6$ & \cellcolor{gray!20}$58.3$ \\
        AKS~\cite{tang2025adaptivekeyframesamplinglong} & 128 & $69.9$ & $72.7$ & $\textbf{67.8}$ & $63.2$ & $65.3$ & $61.5$ & $70.8$ & $\textbf{63.3}$ & \cellcolor{gray!20}$67.0$ & $\textbf{37.0}$ & $\textbf{74.5}$ & $58.3$ & $56.1$ & $51.4$ & $53.4$ & $52.4$ & $53.6$ & $\textbf{76.5}$ & \cellcolor{gray!20}$58.6$ \\
        Q-Frame~\cite{zhang2025qframequeryawareframeselection} & 128 & $65.6$ & $69.3$ & $60.9$ & $59.2$ & $65.3$ & $61.5$ & $67.7$ & $59.5$ & \cellcolor{gray!20}$63.7$ & $35.6$ & $71.3$ & $61.1$ & $48.5$ & $51.4$ & $53.4$ & $52.4$ & $\textbf{60.8}$ & $70.4$ & \cellcolor{gray!20}$56.9$ \\
        \rowcolor{gray!20} \rowcolor{gray!20} \textbf{DIG}~(\textit{Ours}) & 128 & $\textbf{76.3}$ & $\textbf{83.0}$ & $65.5$ & $\textbf{65.8}$ & $\textbf{70.8}$ & $\textbf{67.7}$ & $\textbf{80.0}$ & $\textbf{63.3}$ & $\textbf{71.6}$ & $34.3$ & $72.3$ & $\textbf{68.1}$ & $\textbf{68.2}$ & $\textbf{66.2}$ & $\textbf{57.5}$ & $45.1$ & $55.7$ & $74.1$ & $\textbf{60.2}$ \\
        \midrule
        UNI & 192 & $72.0$ & $73.9$ & $66.7$ & $67.1$ & $66.7$ & $\textbf{67.7}$ & $72.3$ & $63.3$ & \cellcolor{gray!20}$68.8$ & $34.3$ & $\textbf{76.6}$ & $58.3$ & $62.1$ & $58.1$ & $\textbf{60.3}$ & $\textbf{52.4}$ & $\textbf{56.7}$ & $71.6$ & \cellcolor{gray!20}$59.4$ \\
        AKS~\cite{tang2025adaptivekeyframesamplinglong} & 192 & $71.0$ & $79.5$ & $\textbf{67.8}$ & $67.1$ & $66.7$ & $60.0$ & $69.2$ & $62.0$ & \cellcolor{gray!20}$68.3$ & $35.6$ & $74.5$ & $58.3$ & $57.6$ & $54.1$ & $50.7$ & $48.8$ & $54.6$ & $\textbf{76.5}$ & \cellcolor{gray!20}$57.3$ \\
        \rowcolor{gray!20} \rowcolor{gray!20} \textbf{DIG}~(\textit{Ours}) & 192 & $\textbf{73.1}$ & $\textbf{84.1}$ & $\textbf{67.8}$ & $\textbf{68.4}$ & $\textbf{68.1}$ & $64.6$ & $\textbf{76.9}$ & $\textbf{65.8}$ & $\textbf{71.1}$ & $\textbf{38.4}$ & $74.5$ & $\textbf{70.8}$ & $\textbf{69.7}$ & $\textbf{71.6}$ & $57.5$ & $43.9$ & $\textbf{56.7}$ & $75.3$ & $\textbf{60.7}$ \\
        \bottomrule
    \end{tabular}
    }
    \vspace{-9pt}
\end{table}

\subsection{Extended Experiments with DIG}
To investigate the scalability of~\name~in ultra-long context scenarios, we extended our experiments using Qwen3-VL-8B~\cite{bai2025qwen3vltechnicalreport}, an open-source LMM distinguished for its robust long-context processing capability. We test~\name~against the uniform sampling baseline and AKS~\cite{tang2025adaptivekeyframesamplinglong}.

\paragraph{Experiment settings.}
For~\name, the query identification and CAFS configurations align with Section~\ref{sec:experiment}, with the exception that we employ Qwen3-VL-8B~\cite{bai2025qwen3vltechnicalreport} as the unified backbone for both reward assignment and final inference. Similarly, AKS~\cite{tang2025adaptivekeyframesamplinglong} setup mirrors Section~\ref{sec:experiment} but utilizes Qwen3-VL-8B~\cite{bai2025qwen3vltechnicalreport} as the base model. To rigorously test performance across varying context lengths, we scaled input frame counts from 8 to 768, with each frame encoded into approximately 150 tokens. The results are in Table~\ref{tab:qwen3vl}.

\paragraph{DIG delivers consistent performance gains.}
As evidenced in Table~\ref{tab:qwen3vl},~\name~yields substantial improvements across nearly all frame configurations. Notably, with 256 input frames, DIG achieves an $8.6\%$ performance boost on MLVU~\cite{zhou2025mlvubenchmarkingmultitasklong} compared to uniform sampling. Crucially, DIG maintains robustness even at the extreme scale of 768 frames, surpassing the baseline by $4.7\%$ on MLVU~\cite{zhou2025mlvubenchmarkingmultitasklong}, $3.7\%$ on LongVideoBench~\cite{wu2024longvideobenchbenchmarklongcontextinterleaved}, and $3.5\%$ on VideoMME-Medium~\cite{fu2024videommefirstevercomprehensiveevaluation}. In contrast, while AKS~\cite{tang2025adaptivekeyframesamplinglong} remains competitive at lower frame counts ($\le 64$), it exhibits marked performance degradation as the context length increases, frequently falling below the uniform sampling baseline. Given that practical video understanding tasks necessitate maximizing input frames to capture comprehensive temporal details, AKS~\cite{tang2025adaptivekeyframesamplinglong} demonstrates limited utility for real-world applications. Conversely,~\name~exhibits superior scalability, effectively delivering sustained performance gains.

\subsection{Detailed Experiment Results \& More Analysis}
We present detailed performance breakdowns corresponding to the benchmarks discussed in Section~\ref{sec:experiment}. Comprehensive quantitative results are in Tables~\ref{tab:exp_mlvu},~\ref{tab:exp_videomme_detail}, and~\ref{tab:exp_lv_detail}.

\paragraph{Uniform sampling suffices for global queries.}
For global queries, specifically Anomaly Recognition and Topic Reasoning tasks within MLVU~\cite{zhou2025mlvubenchmarkingmultitasklong}, all evaluated methods perform comparably to uniform sampling, regardless of the input frame count. This observation reaffirms our previous assertion: uniform sampling is the preferred strategy for global queries, as it achieves sufficient performance while maintaining high efficiency.

\vspace{6pt}
\noindent Inference for localized queries operates in two distinct stages: query-aware frame selection and subsequent reasoning based on the retrieved content. Without the initial selection stage, evaluating the model's fundamental performance is challenging, as errors may stem from information-deficient inputs rather than inherent model limitations. By incorporating this stage to ensure the input contains relevant information, we can decouple data retrieval issues from reasoning capabilities. This allows for a more accurate assessment of the model's intrinsic proficiency across different tasks, yielding deeper insights.

\paragraph{Query-aware selection uncovers intrinsic visual perception capabilities.}
As shown in Table~\ref{tab:exp_mlvu} and~\ref{tab:exp_lv_detail}, our method significantly and consistently outperforms uniform sampling on localized perception tasks (e.g., PlotQA, NeedleQA, and L1-Perception). Notably, these tasks primarily evaluate fundamental visual perception capabilities. Our findings suggest that LMMs are intrinsically capable of solving such tasks, provided the query-relevant information is effectively supplied. This explains the substantial performance gap: while uniform sampling often introduces significant noise by including irrelevant content, query-aware selection ensures the model is conditioned on relevant frames.

\paragraph{Temporal reasoning remains a fundamental bottleneck.}
Conversely, regarding tasks requiring temporal logic (e.g., Action Order and L2-Relation), performance remains stagnant across all methods. Even when provided with query-relevant visual information, model performance does not improve. This underscores a critical limitation: current LMMs struggle with temporal reasoning and sequencing, a deficiency that persists independently of the quality of visual information retrieval.

\clearpage

%% file: supp/7_time.tex
\section{More Efficiency Analysis of DIG}
\label{sec:time}

\subsection{Detailed Runtime Profiling}
We evaluate the computational efficiency of~\name~compared to distinct baselines, AKS~\cite{tang2025adaptivekeyframesamplinglong} and Q-Frame~\cite{zhang2025qframequeryawareframeselection}. The total runtime of each method can be divided into two stages: 
\begin{itemize}
    \item \textit{Key Frame Selection}, where the method identifies optimal indices from raw video.
    \item \textit{Inference}, where the LMM processes the selected frames to generate a response.
\end{itemize}
\noindent All experiments were conducted on a node equipped with 8 NVIDIA A100 GPUs. To provide a comprehensive analysis, we report the standard LMM inference latency across varying input frame counts in Table~\ref{tab:inference_time} and detail the selection overhead introduced by specific methods in Table~\ref{tab:selection_cost}.

\paragraph{DIG achieves a favorable efficiency-performance trade-off.} 
As evidenced in Table~\ref{tab:selection_cost},~\name~offers a significant efficiency advantage over AKS~\cite{tang2025adaptivekeyframesamplinglong}, reducing computational overhead by an order of magnitude while maintaining superior downstream performance (see Section~\ref{sec:experiment}). While~\name~incurs a marginal increase in processing time compared to Q-Frame~\cite{zhang2025qframequeryawareframeselection}, this cost is justified by substantial robustness gains; specifically, Q-Frame~\cite{zhang2025qframequeryawareframeselection} fails to outperform uniform sampling as frame counts exceed 32, whereas~\name~consistently surpasses baselines across all settings. Furthermore, comparing the selection overhead (Table~\ref{tab:selection_cost}) against standard inference latency (Table~\ref{tab:inference_time}), the additional cost remains within a reasonable range. This confirms that~\name~effectively balances efficiency and accuracy, serving as a practical, plug-and-play module for enhanced long-form video understanding.

\begin{table}[h]
  \centering
  \caption{\textbf{Inference latency analysis.} The inference time (in minutes) of the base LMM (Qwen2.5-VL-7B~\cite{bai2025qwen25vltechnicalreport}) across different input frame counts using standard uniform sampling.}
  \label{tab:inference_time}
  \resizebox{0.5\linewidth}{!}{
    \begin{tabular}{l ccccccc}
    \toprule
       \multirow{2}{*}{Dataset} & \multicolumn{7}{c}{\# Frames} \\
       \cmidrule(lr){2-8}
       & 8 & 16 & 32 & 64 & 128 & 192 & 256 \\
       \midrule
       MLVU~\cite{zhou2025mlvubenchmarkingmultitasklong} & 3.2 & 5.0 & 9.3 & 17.6 & 29.1 & 37.3 & 43.4\\
       LongVideoBench~\cite{wu2024longvideobenchbenchmarklongcontextinterleaved} & 1.4 & 2.2 & 4.3 & 8.3 & 14.0 & 19.9 & 25.6\\
       VideoMME~\cite{fu2024videommefirstevercomprehensiveevaluation} & 3.1 & 4.7 & 8.7 & 15.8 & 26.1 & 36.7 & 46.3\\
       \bottomrule
    \end{tabular}
  }
  \vspace{12pt}
  \caption{\textbf{Comparison of frame selection overhead.} The time cost (in minutes) required by different methods to process videos and select key frames. For DIG, we break down the cost into Query Identification~(QI), Content-Aware Frame Selection~(CAFS), Reward Assignment~(RA), and Video Refinement~(VR).}
  \label{tab:selection_cost}
  \resizebox{0.7\linewidth}{!}{
    \begin{tabular}{l cc ccccc}
    \toprule
       \multirow{2}{*}{Dataset} & \multirow{2}{*}{AKS~\cite{tang2025adaptivekeyframesamplinglong}} & \multirow{2}{*}{Q-Frame~\cite{zhang2025qframequeryawareframeselection}} & \multicolumn{5}{c}{DIG~(Ours)}\\
       \cmidrule(lr){4-8}
       & & & QI & CAFS & RA & VR & Sum \\
       \midrule
       MLVU~\cite{zhou2025mlvubenchmarkingmultitasklong} & $\geq 720$ & 122.1 & 11.3 & 25.9 & 218.9 & 0.2 & 256.3\\
       LongVideoBench~\cite{wu2024longvideobenchbenchmarklongcontextinterleaved} & $\geq 720$ & 34.5 & 7.6 & 20.8 & 110.4 & 0.1 & 138.9\\
       VideoMME~\cite{fu2024videommefirstevercomprehensiveevaluation} & $\geq 720$ & 94.2 & 11.6 & 31.2 & 264.8 & 0.3 & 307.9\\
       \bottomrule
    \end{tabular}
  }
  \vspace{-9pt}
\end{table}

\subsection{Efficiency Gains from Query Identification}
To balance efficiency and accuracy, \name~employs a Query Identification module. We apply resource-intensive key frame selection only to localized queries, defaulting to efficient uniform sampling for global ones. This adaptive strategy minimizes computational cost without compromising downstream performance (see Section~\ref{sec:discussion}). Table~\ref{tab:qi_efficiency} quantifies these gains by comparing our adaptive approach against the baseline that applies our specific selection universally. On LongVideoBench~\cite{wu2024longvideobenchbenchmarklongcontextinterleaved}, where queries are predominantly localized, the QI module incurs a marginal overhead ($3.6\%$) due to the additional classification step. However, on datasets with a diverse mix of query types, such as VideoMME~\cite{fu2024videommefirstevercomprehensiveevaluation} and MLVU~\cite{zhou2025mlvubenchmarkingmultitasklong}, the adaptive strategy yields significant time savings ($19.9\%$ and $13.3\%$, respectively). This demonstrates that the QI module effectively optimizes resource allocation by bypassing unnecessary computation for global queries.

\begin{table}[t]
  \centering
  \small
  \providecommand{\cgreen}[1]{\textcolor[rgb]{0.0, 0.6, 0.0}{#1}} 
  \providecommand{\cred}[1]{\textcolor{red}{#1}}

  \caption{\textbf{Impact of Query Identification on efficiency.} We compare the frame selection time (in minutes) of applying our specific selection universally (w/o. QI) versus DIG's adaptive approach (w. QI). ``Percent'' denotes the proportion of localized queries.}
  \label{tab:qi_efficiency}
  
  \resizebox{0.6\linewidth}{!}{
    \begin{tabular}{l|c|c|r@{\hspace{4pt}}l}
       \toprule
       & Percent & w/o QI & \multicolumn{2}{c}{w/ QI}\\
       \midrule
       MLVU~\cite{zhou2025mlvubenchmarkingmultitasklong} & 82.8 & 295.7 & 256.3 & \cgreen{($\downarrow$ 13.3\%)} \\
       LongVideoBench~\cite{wu2024longvideobenchbenchmarklongcontextinterleaved} & 97.8 & 134.1 & 138.9 & \cred{($\uparrow$ 3.6\%)} \\
       VideoMME~\cite{fu2024videommefirstevercomprehensiveevaluation} & 77.0 & 384.2 & 307.9 & \cgreen{($\downarrow$ 19.9\%)} \\
       \bottomrule
    \end{tabular}
  }
  
\end{table}